\documentclass[conference]{IEEEtran}
\ifCLASSINFOpdf
\else
\fi
\usepackage{cite}
\usepackage{amsmath,amssymb,amsfonts}
\usepackage{algorithmic}
\usepackage{graphicx}
\usepackage{textcomp}
\usepackage{tikz}
\usepackage{pgfplots}

\usepackage{subcaption}

\usepackage{multicol}
\usepackage{siunitx}

\usepackage{setspace}

\usepackage{pifont}
\newcommand\copyrighttext{%
	\scriptsize \copyright~2020 IEEE. Personal use of this material is permitted. Permission from IEEE must be obtained for all other uses, in any current or future media, including reprinting/republishing this material for advertising or promotional purposes, creating new collective works, for resale or redistribution to servers or lists, or reuse of any copyrighted component of this work in other works.}%
\newcommand\copyrightnotice{%
	\begin{tikzpicture}[remember picture,overlay]
	\node[anchor=south,yshift=10pt,xshift=0.25cm] at (current page.south) {{\parbox{\dimexpr\textwidth-\fboxsep-\fboxrule\relax}{\copyrighttext}}};
	\end{tikzpicture}%
}

\begin{document}
	
	\title{Trajectory Planning for Automated Driving in Intersection Scenarios using Driver Models\\
		\thanks{*This work was supported by Visteon Corporation}
	}
	
	\author{
		\IEEEauthorblockN{Oliver Speidel}
		\IEEEauthorblockA{\textit{Measurement-, Control- and Microtech.} \\
			\textit{Ulm University}\\
			89081 Ulm, Germany\\
			oliver.speidel@uni-ulm.de}
		\and
		\IEEEauthorblockN{Maximilian Graf}
		\IEEEauthorblockA{\textit{Measurement-, Control- and Microtech.} \\
			\textit{Ulm University}\\
			89081 Ulm, Germany\\
			maximilian.graf@uni-ulm.de}
		\and
		\IEEEauthorblockN{Ankit Kaushik}
		\IEEEauthorblockA{\textit{ADAS Functions}\\
			\textit{Visteon Electronics GmbH Germany} \\
			76227 Karlsruhe, Germany \\
			ankit.kaushik@visteon.com}
		\and
		\hspace{0.8cm}\IEEEauthorblockN{Thanh Phan-Huu}
		\IEEEauthorblockA{\hspace{0.8cm}\textit{ADAS Functions}\\
			\hspace{0.8cm}	\textit{Visteon Electronics GmbH Germany} \\
			\hspace{0.8cm}	76227 Karlsruhe, Germany \\
			\hspace{0.8cm}	tphanhuu@visteon.com}
		\and
		\IEEEauthorblockN{Andreas Wedel}
		\IEEEauthorblockA{\textit{ADAS Functions}\\
			\textit{Visteon Electronics GmbH Germany} \\
			76227 Karlsruhe, Germany \\
			andreas.wedel@visteon.com}
		\and
		\IEEEauthorblockN{Klaus Dietmayer \hspace{-0.6cm}}
		\IEEEauthorblockA{\textit{Measurement-, Control- and Microtech.\hspace{-0.6cm}} \\
			\textit{Ulm University \hspace{-0.6cm}}\\
			89081 Ulm, Germany \hspace{-0.6cm}\\
			klaus.dietmayer@uni-ulm.de \hspace{-0.6cm}}
	}
	
	\maketitle
	\copyrightnotice
	\thispagestyle{empty}
	\pagestyle{empty}
	\begin{abstract}
		Efficient trajectory planning for urban intersections is currently one of the most challenging tasks for an Autonomous Vehicle (AV). Courteous behavior towards other traffic participants, the AV's comfort and its progression in the environment are the key aspects that determine the performance of trajectory planning algorithms. To capture these aspects, we propose a novel trajectory planning framework that ensures social compliance and simultaneously optimizes the AV's comfort subject to kinematic constraints. 
		The framework combines a local continuous optimization approach and an efficient driver model to ensure fast behavior prediction, maneuver generation and decision making over long horizons.
		The proposed framework is evaluated in different scenarios to demonstrate its capabilities in terms of the resulting trajectories and runtime.
	\end{abstract}
	
	\begin{IEEEkeywords}
		autonomous driving; trajectory planning; optimization; autonomous vehicle; decision making
	\end{IEEEkeywords}
	
	\section{Introduction}
	\subsection{Motivation}
\begin{figure}[t]
		\centering
		\def\svgwidth{0.425\columnwidth}
		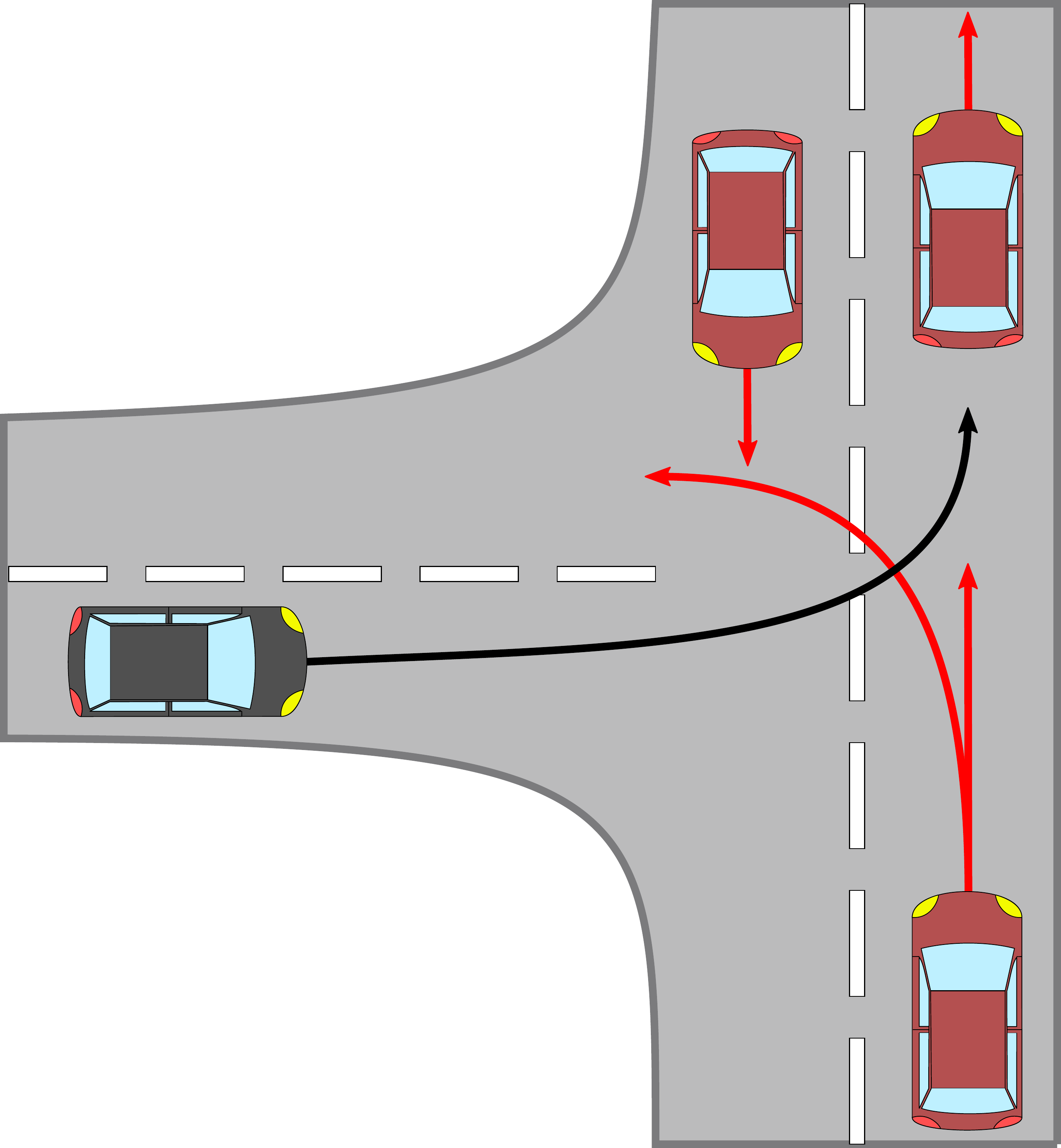
		\caption{Exemplary Intersection Scenario, where multiple vehicles with different maneuver options have to be regarded.}
		\label{fig:overview}
	\end{figure}

	In past few years, the automobile industry is undergoing a drastic change in the driving features related to autonomous driving \cite{Urmson2008,Kunz2015}.
	An Autonomous Vehicle (AV) has to successfully ensure that it is capable of operating in complex scenarios such as urban intersections. As a result, the AV has to perceive the environmental model with high precision and perform trajectory planning in real-time which becomes highly challenging while driving inner-city. Therefore, efficient trajectory planning algorithms that satisfy real-time constraints are absolutely necessary.
	Several approaches \cite{Werling2010,Graf2018,Ziegler2014} came forward that considered comfort, vehicle dynamics, and other vehicles on the ego lane.
	Recently, trajectory planning has been extended to include longer time horizons for more complex scenarios whereby taking other traffic participants at multiple lanes into account \cite{Tas2018,Hoermann2017,Evestedt2016}.
	In this paper, we investigate trajectory planning in complex intersection scenarios as depicted in \figurename~\ref{fig:overview} while preserving comfort and real-time capability.
	A further significant aspect is the social compliance of the AV's behavior. 
	In dense traffic scenarios, too passive behavior might lead to delays in the ego progress when encountered with an intersection. Therefore, a major objective for a trajectory planner is to preserve the AV's progress and at the same time ensure that the performed maneuvers are appropriate to other vehicles.
	In order to handle such complex problem formulations, a popular approach is to use sampling-based solutions that are limited to a discrete action space \cite{Hoermann2017,Evestedt2016}.
	Whereas local continuous approaches are often restricted due to their computation times or are limited to specific \mbox{scenarios \cite{Graf2018,Ziegler2014,Tas2018}}.
	
	In this regard, we propose a planning framework including a local continuous optimization technique that ensures real-time computation and, thereby, increasing its applicability for intersection scenarios while considering social compliance.
	
	\subsection{Related Work}
	Recently, local continuous optimization techniques have often been applied to trajectory planning for autonomous driving \cite{Graf2018, Ziegler2014, Tas2018}.
	In contrast, a common concept for trajectory generation especially in intersection scenarios is to sample jerk optimal polynomials \cite{Werling2010,Hoermann2017}. According to which, multiple trajectory candidates are generated and subsequently checked for potential collisions with the objective of performing a safe maneuver over the intersection. To guarantee reachability to a safe state, the regarded horizon has to be long enough \cite{Hoermann2017}. Therefore, the sampled trajectories might have to be extrapolated with constant acceleration to verify safety \cite{Hoermann2017}. The extrapolation, however, \mbox{limits the applicability for long horizons.} 
	
	Social compliant behavior can be generated by choosing trajectories with low influence on other vehicles behavior. This effect is measured, for example, by predicting the reaction of upcoming vehicles with the Intelligent Driver Model (IDM) \cite{Treiber2000}. The increase in disturbance of upcoming traffic results in the degradation of the social compliance \cite{Evestedt2016, Speidel2019}.
	In addition, the IDM has already been applied for decision making in lane change scenarios \cite{Kesting2007}. This is done by estimating the acceleration response of other vehicles caused by the lane change trajectory of the ego vehicle. The lane change maneuver is executed by the ego vehicle only if the overall costs of the considered vehicles, including the ego vehicle, are improved compared to the current situation.
	Further approaches investigate more sophisticated consideration of interactions in intersection scenarios  using Partially Observable Markov Decision Processes (POMDPs) \cite{Hubmann2018, Brechtel2014}.
	However, these approaches are either limited to a certain type of scenario, are computationally intensive, i.e., not applicable for real-time applications or yield rough discrete actions without any further optimization.
	
	In this regard, a promising approach for real-world application is to combine the advantages of local continuous optimization with efficient driver models in order to enable social compliant car-following and lane change behavior while keeping the computation time low \cite{Graf2018,Graf2019}. A comfortable and drivable trajectory is acquired by generating a reference trajectory using the IDM which is then integrated into a local continuous optimization problem. The lane change decision is evaluated at each time step during replanning using the concept presented in \cite{Kesting2007}.
	
	We extend this approach and additionally embed it into an efficient framework to generate real-time social compliant trajectories for urban intersections over long horizons while considering acceleration constraints.

	\subsection{Overview}
	In this paper, we present a novel trajectory planning framework. The significant aspects of which are summarized as follows, we first process the environment information to procure map data and predictions of other vehicles.
	Afterwards, multiple behavior options for the ego vehicle are generated which are represented as reference trajectories.
	During decision making, the best behavior option is chosen that ensures social compliance by considering the costs arising for other vehicles due to the planned behavior option.
	In general, the behavior options generation and the decision making are based on the same longitudinal car-following driver model \cite{Treiber2000}.
	For this purpose, a strategy is presented to create behavior options in intersection scenarios utilizing a longitudinal car-following driver model.
	Finally, we integrate the reference trajectory with lowest costs into a local continuous optimization problem which regards longitudinal and lateral movement combined subject to constraints on vehicle kinematics.
	
	To summarize the above discussion, the key contributions of this work are:
	\begin{itemize}
		\item the holistic overall framework which extends the concepts presented in \cite{Graf2018,Graf2019} for social compliant behavior in intersection scenarios
		\item a novel concept for behavior generation in intersection scenarios utilizing driver models
		\item a new decision making approach considering the trade-off between ego progress  and costs of other vehicles
		\item the utilization of a new, efficient prediction concept for the consideration of other vehicles behavior options
	\end{itemize}
	In general, the resulting framework performs real-time planning over long horizons while considering other vehicles at urban intersections. Furthermore, the concept yields social compliant maneuvers while preserving the progress of the ego vehicle.
	
	\begin{figure}
		\centering
		\def\svgwidth{\columnwidth}
		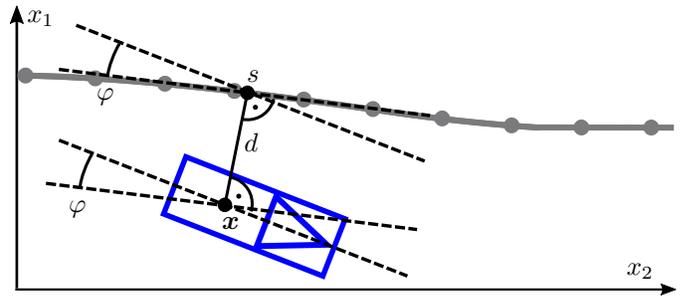
		\caption{Environment Representation, the position in Cartesian coordinates is represented by $\boldsymbol{x}$ and the 1-dimensional position on the center line is denoted by $s$. The lateral deviation of the vehicle to the center line is given with $d$. The difference between the orientation of the center line at $s$ and the orientation of the vehicle is denoted by $\varphi$. The center line is represented by a polygon line. } 
		\label{fig:envModel}
	\end{figure}
	\begin{figure}[]
		\begin{subfigure}[c]{0.4\columnwidth}
			\centering
			\def\svgwidth{\columnwidth}
			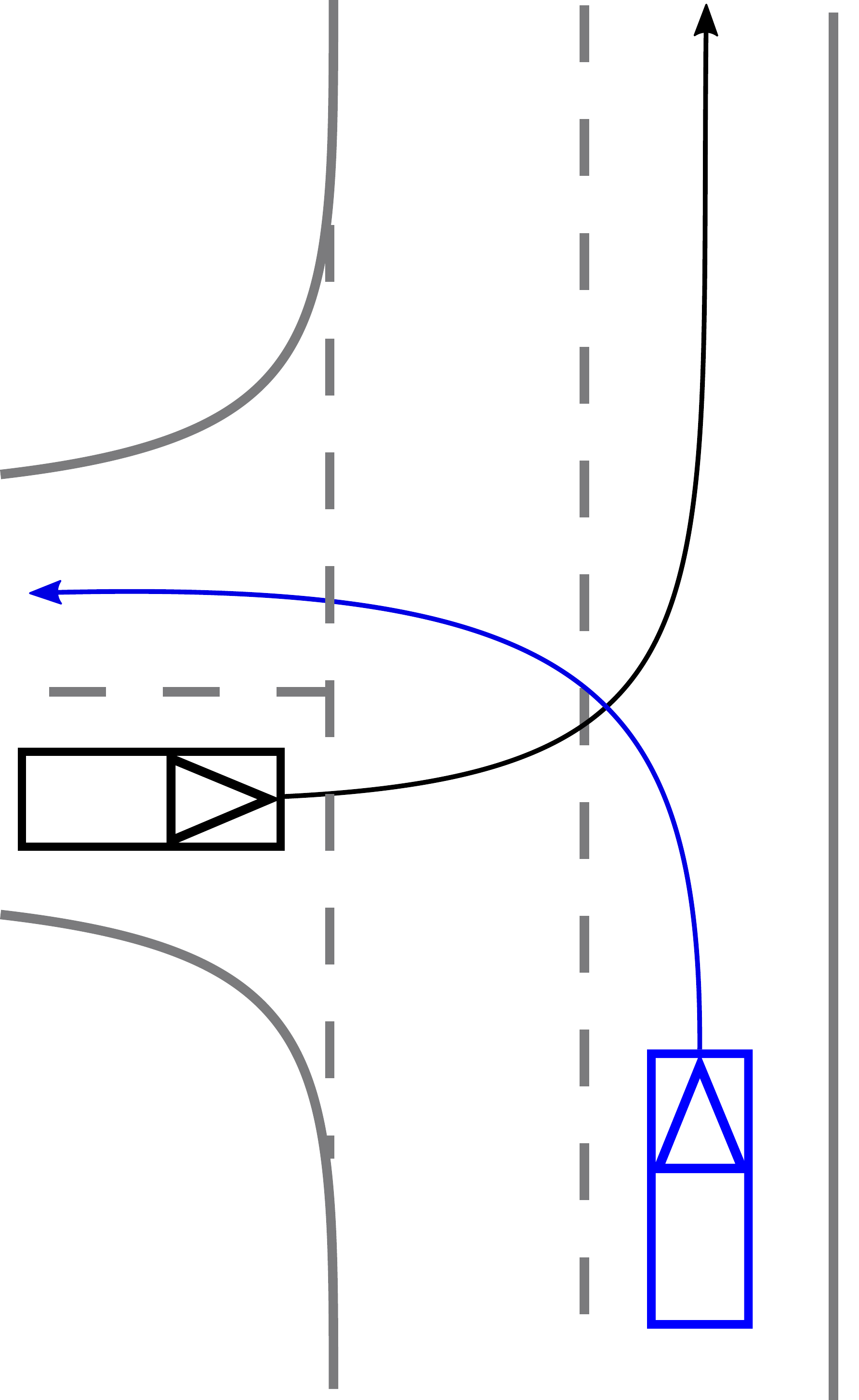
			\caption{Crossing Routes}
			\label{fig:conflictZonesCrossing}
		\end{subfigure}
		\begin{subfigure}[c]{0.4\columnwidth}
			\centering
			\def\svgwidth{\columnwidth}
\begingroup%
  \makeatletter%
  \providecommand\color[2][]{%
    \errmessage{(Inkscape) Color is used for the text in Inkscape, but the package 'color.sty' is not loaded}%
    \renewcommand\color[2][]{}%
  }%
  \providecommand\transparent[1]{%
    \errmessage{(Inkscape) Transparency is used (non-zero) for the text in Inkscape, but the package 'transparent.sty' is not loaded}%
    \renewcommand\transparent[1]{}%
  }%
  \providecommand\rotatebox[2]{#2}%
  \newcommand*\fsize{\dimexpr\f@size pt\relax}%
  \newcommand*\lineheight[1]{\fontsize{\fsize}{#1\fsize}\selectfont}%
  \ifx\svgwidth\undefined%
    \setlength{\unitlength}{513.14007134bp}%
    \ifx\svgscale\undefined%
      \relax%
    \else%
      \setlength{\unitlength}{\unitlength * \real{\svgscale}}%
    \fi%
  \else%
    \setlength{\unitlength}{\svgwidth}%
  \fi%
  \global\let\svgwidth\undefined%
  \global\let\svgscale\undefined%
  \makeatother%
  \begin{picture}(1,1.63071602)%
    \lineheight{1}%
    \setlength\tabcolsep{0pt}%
    \put(0,0){\includegraphics[width=\unitlength,page=1]{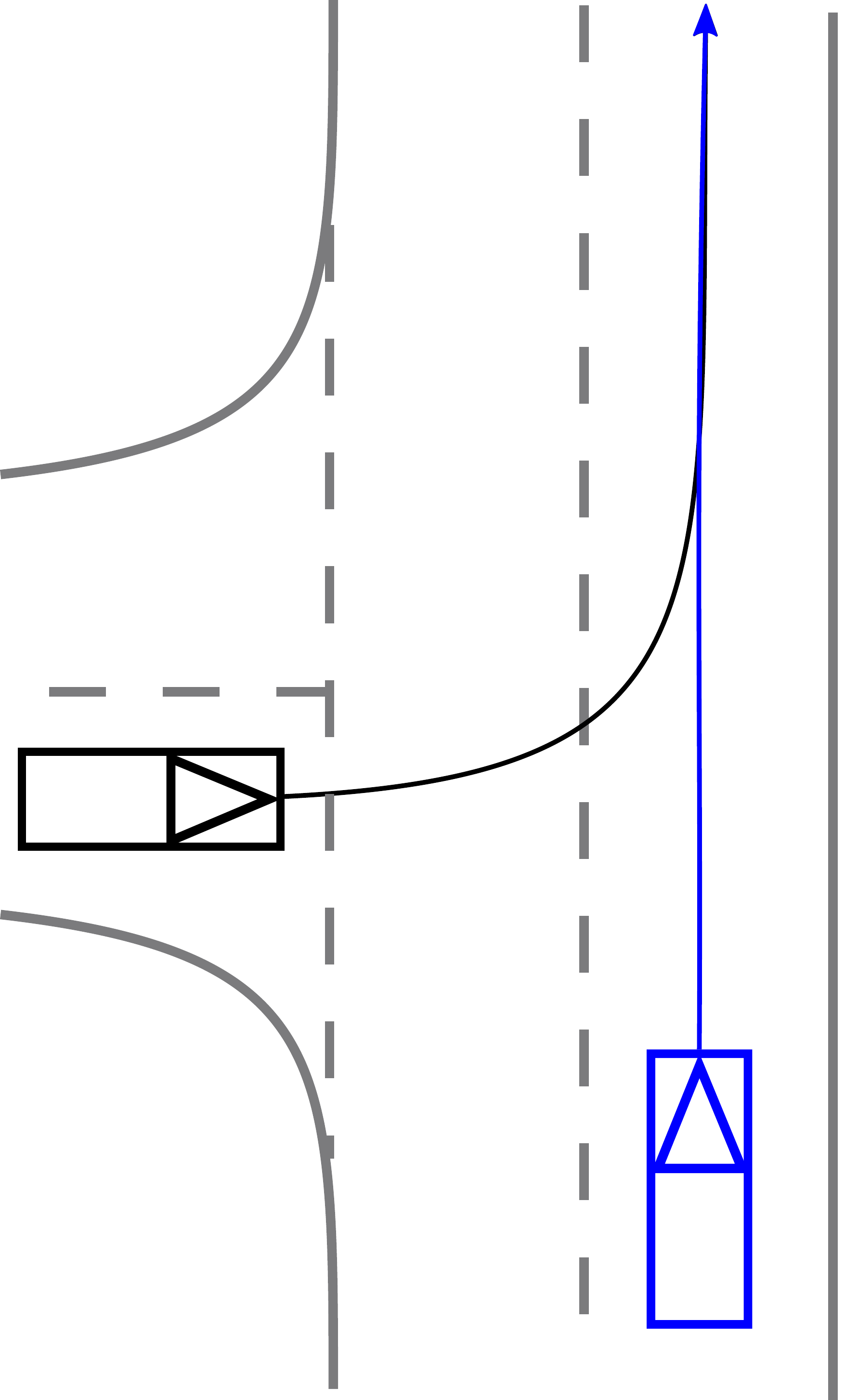}}%
    \put(0.54065999,0.65438325){\color[rgb]{0,0,0}\makebox(0,0)[lt]{\lineheight{1.25}\smash{\begin{tabular}[t]{l}$s^{m}_{a,c}$\end{tabular}}}}%
    \put(0.39939979,0.64360796){\color[rgb]{0,0,0}\makebox(0,0)[lt]{\lineheight{1.25}\smash{\begin{tabular}[t]{l}$s^{s}_{a}$\end{tabular}}}}%
    \put(0,0){\includegraphics[width=\unitlength,page=2]{collision_zones_2_merging_draft.pdf}}%
    \put(0.83921117,0.48059925){\color[rgb]{0,0,0}\makebox(0,0)[lt]{\lineheight{1.25}\smash{\begin{tabular}[t]{l}$r_c$\end{tabular}}}}%
    \put(0.42522979,0.76855515){\color[rgb]{0,0,0}\makebox(0,0)[lt]{\lineheight{1.25}\smash{\begin{tabular}[t]{l}$r_a$\end{tabular}}}}%
    \put(0,0){\includegraphics[width=\unitlength,page=3]{collision_zones_2_merging_draft.pdf}}%
    \put(0.82370225,0.57133127){\color[rgb]{0,0,0}\makebox(0,0)[lt]{\lineheight{1.25}\smash{\begin{tabular}[t]{l}$s^{m}_{c,a}$\end{tabular}}}}%
  \end{picture}%
\endgroup%

			\caption{Merging Routes}
			\label{fig:conflictZonesMerging}
		\end{subfigure}
		\caption{Conflict Zones. (a) shows crossing routes with the conflict positions for route $r_a$ and $r_b$. The start of the conflict zone is denoted by $s^{cs}$ and the end by $s^{ce}$. The start of an intersection for a route may differ from the start of a conflict zone and is denoted by $s^s$. (b) depicts a merging scenario where the conflict zone only has a start which is represented by $s^m$. After $s^m$ both vehicles share the same lane.}
		\label{fig:conflictZones}
	\end{figure}
	
	\section{Trajectory Planning Framework}
	\subsection{Environment Representation}
	\label{subsec:EnvironmentModel}
	The environment is represented in Cartesian coordinates. The single lanes of the roads are represented by their center lines which are stored as polygon lines. In addition, the position of a vehicle's gravity center is described by \mbox{$\boldsymbol{x} = [x_1,x_2]^{\rm T}$}.
	The vehicle states can be projected onto a center line to obtain a 1-dimensional longitudinal position $s$, the lateral deviation to the center line $d$ and the orientation difference $\varphi$, using the method described in \cite{Ziegler2014}. The absolute velocity of a vehicle is denoted by $v$.
	The notations are summarized in \figurename~\ref{fig:envModel}.
	The road structure in intersections is represented by routes $r$, which are concatenations of lanes entering and leaving the intersection. Consequently, a route represents a complete path through the intersection. As a result, routes may cross or merge into one another. A conflict zone is described by the longitudinal sections of two routes where they overlap each other. 
	Conflict zones are defined for crossing and merging routes. 
	The detailed calculation of the conflict zone is of minor importance and therefore omitted in this work. The conflict zone for a route $r_{a}$ which crosses $r_{b}$ is given by the start position $s^{cs}_{a,b}$ and end position $s^{ce}_{a,b}$. In contrast, the conflict zone of a route $r_a$ merging with a route $r_c$ is only represented by a start position $s^{sm}_{a,c}$ as up from this position the same lane is shared, cf.  \figurename~\ref{fig:conflictZones}.
	This formulation of the environment in longitudinal coordinates along center lines enables us to deploy longitudinal driver models.
	
	\subsection{Intelligent Driver Model}
	The presented trajectory planning framework is kept generic for any longitudinal car-following driver model. However, the IDM is a powerful approach to describe human behavior \cite{Treiber2000}. Therefore, we include the IDM into the framework which is described as follows
	\begin{eqnarray}
	\small
	&&\dot{v}_{i} =   a_\text{IDM}  \left[ 1-\left( \frac{v_{i}}{v_{0}} \right)^\delta   
	-\left(\frac{s^*(v_{i},\Delta v_{i})}{\Delta s_i}\right)^2 \right]  
	\label{eq:IDM_dyn}
	\\ 
	&&s^*(v_{i},\Delta v_{i})= s_0+v_{i} T+\frac{v_{i}\Delta v_{i}}{2\sqrt{a_\text{IDM}b}}\,.
	\label{eq:IDM_des_dst}
	\end{eqnarray}
	$v_i$ is the current velocity, $\Delta v_i$ the velocity difference to the leading vehicle and $\Delta s_i$ denotes the spatial distance between the front of the regarded vehicle $i$ and the rear of the leading vehicle. The maximum acceleration $a_\text{IDM}$, the comfortable deceleration factor $b$ and the acceleration exponent $\delta$ are constant parameters. The desired distance $s^*$ to the leading vehicle also includes a stand still distance $s_0$ and a time gap $T$.   
	The target speed $v_0$ can be adapted according to the curvature of the road by generating a speed profile along a given route \cite{Liebner2013}.
	For further details it is referred to \cite{Treiber2000} and \cite{Liebner2013}.

	\begin{figure*}
		\centering
		\def\svgwidth{\textwidth}
		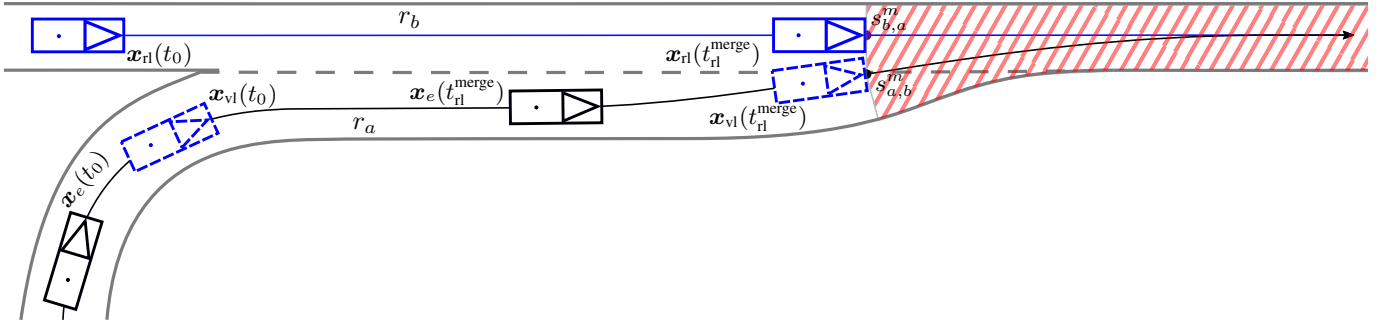
		\caption{Schematic Merging Scenario, where the ego vehicle, the real leading vehicle (RL) and a modeled virtual leading vehicle (VL) are shown. The RL is illustrated with blue solid lanes and the VL with blue dashed lanes. The ego vehicle is drawn in black.
			The ego vehicle approaches on $r_a$ and merges onto $r_b$ on which the RL is driving. The time $t_0$ represents the current time and $t_\text{rl}^\text{merge}$ the point in time when the RL is predicted to enter the shared lane.}
		\label{fig:virtualLeader}
	\end{figure*}
	
	\subsection{Prediction of other vehicles}
	As a part of our framework, we incorporate a longitudinal driver model, specifically the IDM, for predicting other vehicles.
	The goal is to get rough predictions of their future behavior efficiently which can be subsequently considered during planning. Therefore, a set of predicted longitudinal trajectories $\mathcal{T}^{k}_i$ is obtained for each other vehicle $i \in O$, where $ k \in K$ and a trajectory is represented by discrete states $\mathcal{T} = [\boldsymbol{x}^\text{lon}_{0},...,\boldsymbol{x}^\text{lon}_{N_\text{ref}-1}]$ and $\boldsymbol{x}^\text{lon} = [s,v]^{\rm T}$. The time interval between the states is denoted as $ \Delta t$. Further, it is required to estimate the corresponding probability distribution $P(\mathcal{T}|X)$. The environment configuration $X$ contains the position $\boldsymbol{x}$, the velocity $v$, the orientation in Cartesian coordinates of each vehicle and map data.
	Now, according to the IDM, a predicted trajectory depends on the current vehicle state $\boldsymbol{x}^\text{lon}$, a given route $r$ and the prediction of the leading vehicle's trajectory $\mathcal{T}_l$ projected on $r$. Whereby, $\boldsymbol{x}^\text{lon}$ can be directly calculate form $X$ using the projection presented in Section \mbox{\ref{subsec:EnvironmentModel}}. Besides this, we model the maneuver \mbox{$m \in \{\textit{stop, drive}\}$}, where \textit{drive} signifies the ego vehicle passing through the intersection and \textit{stop} means the ego vehicle comes to a stand still at the start of the intersection. Thus, a trajectory $\mathcal{T}_i^{k}$ can be calculated by forward integrating the IDM given $r_i^k$, $m_i^k$, $\mathcal{T}^k_{l_i}$ and $X$. 
	The corresponding probability distribution can be represented by
	\begin{eqnarray}
	\small
	P(\mathcal{T}_i^k|X) &=& P(r_i^k,m_i^k,\mathcal{T}_{l_i}^k|X)\,,
	\end{eqnarray}
	which can be reformulated as 
	\begin{eqnarray}
	\label{eqn:TrajProbDevided}
	\small
	P(\mathcal{T}_i^k|X) &=&  P(r_i^k|X) \cdot P(m_i^k,{T}_{l_i}^k |r_i^k,X)\,.
	\end{eqnarray}
	The term $P(r_i^k|X)$ represents the probability that a vehicle takes the route $r_i^k$. 
	Therefore, a Bayes classifier is employed. To keep the independence of the maneuver in contrast to \cite{Hubmann2017},
	the orientation difference $\varphi^r_i$ of the vehicle to the route $r$ and the corresponding lateral deviation $d^r_i$ are employed as features, cf.~\figurename~\ref{fig:envModel}. In the following, the abbreviations $d^k_i = d^{r_i^k}_i$ and $\varphi^k_i = \varphi^{r_i^k}_i$ are used.
	In general, $\varphi$ and $d$ can be directly extracted from $X$.
	Therefore, it applies
	\begin{equation}
	\small
	\label{eqn:routeProb1}
	P(r_i^k|X) = P(r_i^k|\varphi^k_i, d^k_i) = \frac{P(r_i^k)P(\varphi^k_i, d^k_i|r_i^k)}{P(\varphi^k_i, d^k_i)}\,.
	\end{equation}
	Based on the assumption that $d_i$ and $\varphi_i$ are independent and that every route has the same a-priori probability, $P(r_i^k|X)$ can be formulated as
	\begin{equation}
	\small
	P(r_i^k|X) = \frac{P(\varphi^k_i|r_i^k) P(d^k_i|r_i^k)}{\sum_{r \in\mathcal{R}} P(\varphi^r_i|r) P(d^r_i|r)}\,.
	\end{equation}
	The feature probabilities are modeled with Gaussian distributions such that \mbox{$P(\varphi^r|r) = \mathcal{N}(0,\SI{0.1}{\radian})$} and \mbox{$P(d^r|r) = \mathcal{N}(0,\SI{0.1}{\meter})$}. As a result, each vehicle obtains a probability for each route. 
	Further, the remaining term $P(m_i^k,{T}_{l_i}^k |r_i^k,X)$ of (\ref{eqn:TrajProbDevided}) has to be regarded. It is computationally inefficient to consider all possible leader trajectories and all resulting predictions. 
	Therefore, only the closest leading vehicle with $P(r_i^k)>\delta_r$ and its most likely trajectory following the given route $r_i^k$ is considered. As a result, a single leading trajectory is regarded and instead of the joined distribution $ P(m_i^k,\mathcal{T}_{l_i}^k |r_i^k,X)$, only the conditioned maneuver probability $P(m_i^k|r_i^k,\mathcal{T}_{l_i}^k,X)$ is considered. Consequently, the results have to be normalized in order that \mbox{$\sum_{k \in K} P(\mathcal{T}_i^{k}|X) = 1$}. 
	The maneuver probability $P(m_i^k|r_i^k,\mathcal{T}_{l_i}^k,X)$ is estimated by evaluating if there might be vehicles with right of way which force the regarded vehicle to stop at the intersection. This is done using map information if there are even any other routes with right of way over the regarded route $r_i^k$, expressed by \mbox{$\text{rw}: r \mapsto \{0,1\}$}. In addition, it is considered if there are any actual predictions on one of that prioritized routes which occupy the intersection and might force the regarded vehicle to stop, denoted by \mbox{$\text{I}: (r,\mathcal{T}) \mapsto \{0,1\}$}. It applies $\text{I}(r_i^k,\mathcal{T}_{l_i}^k,X) = 1$ if there is a prioritized vehicle predicted to be in the intersection with a time to collision (TTC) $< \Delta t_\text{inter}$ relative to the regarded vehicle using the \textit{drive} maneuver.
	Hence, the maneuver probability is formulated as
	\begin{equation}
	\small
	P(m_i^k=\textit{drive}|r_i^k,\mathcal{T}_{l_i}^k,X) = \left\{
	\begin{array}{ll}
	\lambda_1& \text{rw}(r_i^k) \land \text{I}(r_i^k,\mathcal{T}_{l_i}^k,X) = 1 \\
	\lambda_2& \text{rw}(r_i^k) \land \text{I}(r_i^k,\mathcal{T}_{l_i}^k,X) = 0 \\
	\lambda_3& \, \textrm{otherwise} \\
	\end{array}
	\right.
	\end{equation}
	and 
	$P(m=\textit{stop}) = 1-P(m=\textit{drive})$.
	The parameters used for this prediction, including $\Delta t_\text{inter} = \SI{3}{\second}$, $\lambda_1  = 0.55$, $\lambda_2 = 1$ and $\lambda_3 = 0.75$, are determined based on heuristics. 
	Alternatively, it is possible to learn the probability distribution \mbox{$P(m^k|r^k,\mathcal{T}_l^k,X)$} or corresponding parameters using measurement data.
	However, in this work, we limit our scope to the former approach, where rough predictions can be generated with low computational effort.
	The evaluation in Section \ref{sec:evaluation} emphasizes the effectiveness of the concept and shows that it enables the framework to choose correct behavior options.

	\subsection{Reference Trajectory Generation}
	\label{subsec:referenceTraj}
	
	\begin{figure}
		\def\svgwidth{0.93\columnwidth}
		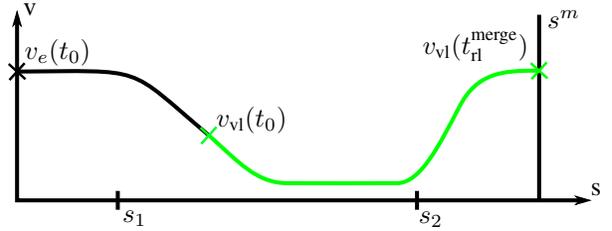
		\caption{Schematic Speed Profile, where the speed profile $v^{\text{idm}}(s)$ beginning at the ego state is depicted in black and the actual speed profile of the virtual leading vehicle (VL) is shown in green. Between $s_1$ and $s_2$ the target velocity is restricted by the curvature of the road.}
		\label{fig:virtualTrajectory}
	\end{figure}
	
	The objective of the reference trajectory generation is to plan multiple behavior options for the ego vehicle. A reference trajectory $\mathcal{T}^{\text{ref}}$ is generated by forward integrating the IDM \cite{Graf2018,Graf2019}.
	Depending on the situation, multiple behavior options or respectively reference trajectories $\{\mathcal{T}^{\text{ref},1},...,  \mathcal{T}^{\text{ref},B}\}$ can be generated. Where $B$ is the number of available behavior options.
	In \cite{Graf2018} and \cite{Graf2019}, the investigation was limited to a single reference trajectory, which was generated based on the prediction of a leading vehicle on the ego lane. However, this approach is not viable for intersection scenarios due to multiple other vehicles on different routes which have to be considered. 
	As a part of the proposed framework, free drive or car-following trajectories regarding vehicles on the ego route, stop trajectories for stopping in front of intersections and merging trajectories for merging behind vehicles of other routes are generated based on the IDM.
	As vehicles of other routes can't be directly regarded in the IDM, we introduce the concept of a virtual leading vehicle (VL) on the ego route which represents a real leading vehicle (RL) on another route. 
	This VL can be regarded in the IDM and enables the ego vehicle to perform a smooth merging maneuver behind the RL, where the RL's most probable trajectory is considered, cf. \figurename~\ref{fig:virtualLeader}.
	To enable smooth car-following behavior, it is reasonable to model the VL also using an IDM trajectory. Now, to ensure continuity between the VL and the RL, the VL and RL should enter the shared lane at the same time $t_\text{rl}^\text{merge}$ and speed.
	One simple approach could be to set the VL onto the ego route at the same distance to the shared lane entry as the RL on the other route. As a result, the same longitudinal trajectory could be applied for the VL as predicted for the RL. However, due to different speed limits and curvatures of the routes, this is not practicable.
	To overcome this limitation a speed profile $v^{\text{idm}}(s)$ is generated on the ego route by forward integrating the ego vehicle using the IDM only considering vehicles on the ego route. The velocity of the RL at the time of entry into the shared lane $t_\text{rl}^\text{merge}$ is used as the target speed and is additionally adapted to the curvature of the ego route. This allows us to model the VL's trajectory by backward integrating on the speed profile beginning at the entry into the shared lane $s^m$ at time $t_\text{rl}^\text{merge}$ up to the current time $t_0$, cf. \figurename~\ref{fig:virtualTrajectory}. 
	Using $v^{\text{idm}}(s)$, the VL states are determined as
	\begin{equation}
	\small
	\boldsymbol{x}^\text{lon}_\text{vl}(t-\Delta t) = \begin{pmatrix}s_{\text{vl}}(t) - v^{\text{idm}}(s_{\text{vl}}(t))\Delta t\\v^{\text{idm}}(s_{\text{vl}}(t))\end{pmatrix}\,. 
	\end{equation}
	Given the VL states, a reference trajectory can be generated for the ego vehicle by using the VL as leading vehicle, which results in a smooth merging trajectory behind the RL.
	Consequently, a behavior option for stopping and multiple options for merging into different traffic gaps are considered when encountered with an intersection. Next, we evaluate the obtained reference trajectories $\{\mathcal{T}^{\text{ref},1},...,  \mathcal{T}^{\text{ref},B}\}$ to decide for the best one.
	
	\subsection{Decision Making}
	The costs for a single ego reference trajectory are defined as
	\begin{equation}
	\small
	c = w_p c_p + w_c \sum_{i\in O} \sum_{k\in K} P(\mathcal{T}^{k}_{i}) c^{k}_{c,i}\,,
	\label{eqn:costfunction}
	\end{equation}
	where $w_p$ and $w_c$ are cost weights. The progress costs for the ego vehicle are denoted by $c_p$ and the courtesy costs $c^k_{c,i}$ represent the social compliance of the regarded reference trajectory to the $i$-th vehicle taking its $k$-th predicted trajectory ($\mathcal{T}^{k}_{i}$).
	$c^k_{c,i}$ is further weighted with $P(\mathcal{T}^{k}_{i})$. In this way, we account for all predicted trajectories of all the merging as well as crossing vehicles.
	Consequently, we effectively incorporate the costs of other vehicles caused by the ego trajectory in the framework. 
	In previous work on lane change scenarios, the trade-off between ego progress and costs for other vehicles caused by the ego vehicle is considered by the planned ego acceleration and corresponding acceleration reactions of other vehicles \cite{Kesting2007}. We extend this concept for the assessment of complete trajectories and for the applicability in intersection scenarios to estimate $c_p$ and $c^{k}_{c,i}$.  
	
	The ego stopping behavior is considered to benchmark the progress and courtesy since it is the most passive trajectory with the minimum progress and maximum courtesy. The progress costs of a reference trajectory are determined as 
	\begin{equation}
	\small
	c_p = \frac{1}{N_\text{ref}-1}\sum_{n=0}^{N_\text{ref}-2} a^\text{stop}_{n} - a_{n}\,,
	\end{equation}
	where $a_{n}$ is the acceleration of the $n$-th discrete state of the regarded reference trajectory and $a^\text{stop}_{n}$ is the corresponding acceleration of the stopping reference trajectory.
	The acceleration values are calculated by finite differences and the number of discrete points in $\mathcal{T}^\text{ref}$ is given by $N_\text{ref}$.
	To reduce calculation time, only trajectories where $a_1 > a^\text{stop}_1$ are considered.
	Following the argumentation in \cite{Speidel2019}, the merging and the crossing scenarios are dealt separately to estimate the courtesy costs.
	\subsubsection{Merging Routes}
	When merging into another lane the reaction of upcoming vehicles has to be regarded. This reaction can be approximated with the IDM by using the ego vehicle as leading vehicle from the moment the ego vehicle is planned to enter the intersection. The courtesy costs are given by
	\begin{equation}
	\small
	\label{eqn:MergingCosts}
	c_{c,i}^{k} = \left\{
	\begin{array}{ll}
	\Delta \bar{a}_{i}^{k}-H_a & \Delta \bar{a}_{i}^{k}-H_a \leq \Delta a_\text{max}\\
	\infty & \, \textrm{otherwise} \\
	\end{array}
	\right.\\,
	\end{equation}
	where $\Delta a_\text{max}$ is the maximum allowed disturbance of other vehicles and $\Delta \bar{a}_{i}^{k}$ represents the predicted acceleration change due to the ego trajectory. The hysteresis value \mbox{$H_a \in \{- H_a^\text{const},H_a^\text{const}\}$} is $- H_a^\text{const}$ if the currently regarded reference trajectory represents the same behavior option as executed in the last planning step and $H_a^\text{const}$ otherwise. Thus, $H_a$ reduces courtesy costs if the behavior option stays the same for subsequent planning cycles. 
	This is reasonable as the other vehicles also try to predict the ego vehicle and react accordingly.
	Further, $\Delta \bar{a}_{i}^{k}$ is defined as
	\begin{equation}
	\small
	\Delta \bar{a}_{i}^{k} = \frac{1}{N_\text{ref}-1}\sum_{n = 0}^{N_\text{ref}-2} |a_{i,n}^{k} - a_{i,n}^{k,\text{react}}|\,,
	\end{equation}
	where $a_{i,n}^{k}$ is the acceleration of the $n$-th state in $\mathcal{T}^{k}_{i}$, i.e., the prediction without influence of the ego vehicle.
	$a_{i,n}^{k,\text{react}}$ is the corresponding acceleration if vehicle $i$ reacts to the ego vehicle 
	using the currently evaluated ego reference trajectory. 
	
	\subsubsection{Crossing Routes}
	Regarding crossing routes, it is not reasonable to approximate the interaction with the IDM, because the vehicles can not be modeled as leader and follower. 
	For this reason, it is argued that there is no reaction of the crossing vehicle expected if there is enough temporal distance between the point in times where the ego vehicle leaves the conflict zone $t^{\text{exit}}$ and the crossing vehicle enters it $t^{\text{entry},k}_{i}$.
	Consequently, $\Delta t_{\text{ttc},i}^k = t^{\text{exit}} - t^{\text{entry},k}_{i}$ is employed.
	With this, the courtesy costs for vehicles on crossing routes are defined as
	\begin{equation}
	\small
	c_{c,i}^{k} = \left\{
	\begin{array}{ll}
	0 & \Delta t_{\text{ttc},i}^k -H_\text{ttc} \geq \Delta t_\text{max} \\
	\infty & \, \textrm{otherwise} \\
	\end{array}
	\right.\\,
	\end{equation}
	where $\Delta t_\text{max}$ is the minimum allowed temporal distance.
	Analog to $H_a$ in (\ref{eqn:MergingCosts}), \mbox{$H_\text{ttc} \in \{-H_\text{ttc}^\text{const}, H_\text{ttc}^\text{const}\}$} equals $-H_\text{ttc}^\text{const}$ if the behavior option stays the same and $H_\text{ttc}^\text{const}$ otherwise.
	Finally, the reference trajectory which optimizes ego progress and social compliance the most can be determined. Subsequently, we employ this reference trajectory to generate an executable trajectory in Cartesian coordinates using a local continuous optimization approach.
	
	\subsection{Optimization Problem}
	\label{subsec:optimization}
	At first, the longitudinal reference trajectory is transformed into Cartesian coordinates to procure the positions $[\boldsymbol{x}_0^\text{ref},..., \boldsymbol{x}_{N_\text{opt}-1}^\text{ref}]$ of the reference trajectory.
	$N_\text{opt}$ represents the number of trajectory support points, or respectively states, which are used for the trajectory generation during the optimization. The time interval $\Delta t$ between the trajectory support points stays the same. 
	It is worthy to note that the horizon for the behavior generation and the generation of the optimized trajectory is explicitly chosen to be different. 
	This is hugely favorable, as the behavior has to be regarded on long horizons, and is therefore generated with computationally efficient driver models. Whereas the computationally more complex optimization problem, which regards comfort and kinematic constraints, can be restricted to a shorter horizon. Moreover, the optimization in Cartesian coordinates allows us to plan trajectories deviating from the center line.
	
	With this, the local continuous optimization problem is formulated as
	\begin{equation}
	\small
	\begin{aligned}
	&\min_{\boldsymbol{x}_4,\boldsymbol{x}_5,\dots,\boldsymbol{x}_{N_\text{opt}-1}}& J(\boldsymbol{x}_4,\boldsymbol{x}_5,\dots,\boldsymbol{x}_{N_\text{opt}-1})& \\
	&\text{s.t.}& h_n(\boldsymbol{x}_{n-1},\boldsymbol{x}_{n},\boldsymbol{x}_{n+1}) \leq 0&, \ n=2,\dots,N_\text{opt}-2\,,
	\end{aligned}
	\end{equation}
	where $J$ is the cost function and $h_n$ defines acceleration constraints.
	The first four positions $\boldsymbol{x}_0,\boldsymbol{x}_1,\boldsymbol{x}_2,\boldsymbol{x}_3$ need to be fixed as derivatives up to $\boldsymbol{x}^{(4)}$ are calculated by finite differences. The cost function is given as
	\begin{equation}
	\small
	\begin{aligned}
	J = &w_{\text{spatial}} \sum_{n=4}^{N_\text{opt}-1} j_{\text{spatial},n} &+& w_{\text{acc}}\sum_{n=4}^{N_\text{opt}-2} j_{\text{acc},n}  \\
	+&w_{\text{jerk}} \sum_{n=4}^{N_\text{opt}-2} j_{\text{jerk},n} &+& w_{\text{snap}} \sum_{n=4}^{N_\text{opt}-3}j_{\text{snap},n}\,,
	\end{aligned}
	\end{equation}
	where $w_{\text{spatial}}, w_{\text{acc}}, w_{\text{jerk}}, w_{\text{snap}}$ denote the weights for the corresponding cost terms. The summands $j_{\text{acc},n}$, $j_{\text{jerk},n}$, $j_{\text{snap},n}$ optimize comfort by penalizing the acceleration, jerk and snap.
	The obtained reference trajectory is integrated into the optimization by the costs $j_{\text{spatial},n}$ which penalize the spatial difference to the single reference trajectory states with
	\begin{eqnarray}
	\small
	j_{\text{spatial},n}&=&||\boldsymbol{x}^\text{ref}_{n}-\boldsymbol{x}_n||_2^2 \,.
	\end{eqnarray}
	For further details, the reader is advised to refer to our previous work \cite{Graf2018}.
	The cost function is subject to a constraint on the absolute acceleration given by \cite{Ziegler2014}
	\begin{equation}
	\small
	h_n(\boldsymbol{x}_{n-1},\boldsymbol{x}_n,\boldsymbol{x}_{n+1})= ||a^\text{abs}_n||_2^2 -a_{\text{max}}^2 \leq 0, \hspace{0.2cm} n=4,\dots,N_\text{opt}-2\,,
	\label{eq:constraints}
	\end{equation}
	where $a^\text{abs}_n$ is the absolute acceleration of the $n$-th trajectory state calculated by finite differences and $a_{\text{max}}$ is the maximum allowed absolute acceleration. As a result, a social compliant trajectory with optimized comfort and satisfied acceleration constraints is obtained by utilizing the best behavior option.
	
	\section{Evaluation}
	\label{sec:evaluation}
	\begin{figure*}[t]
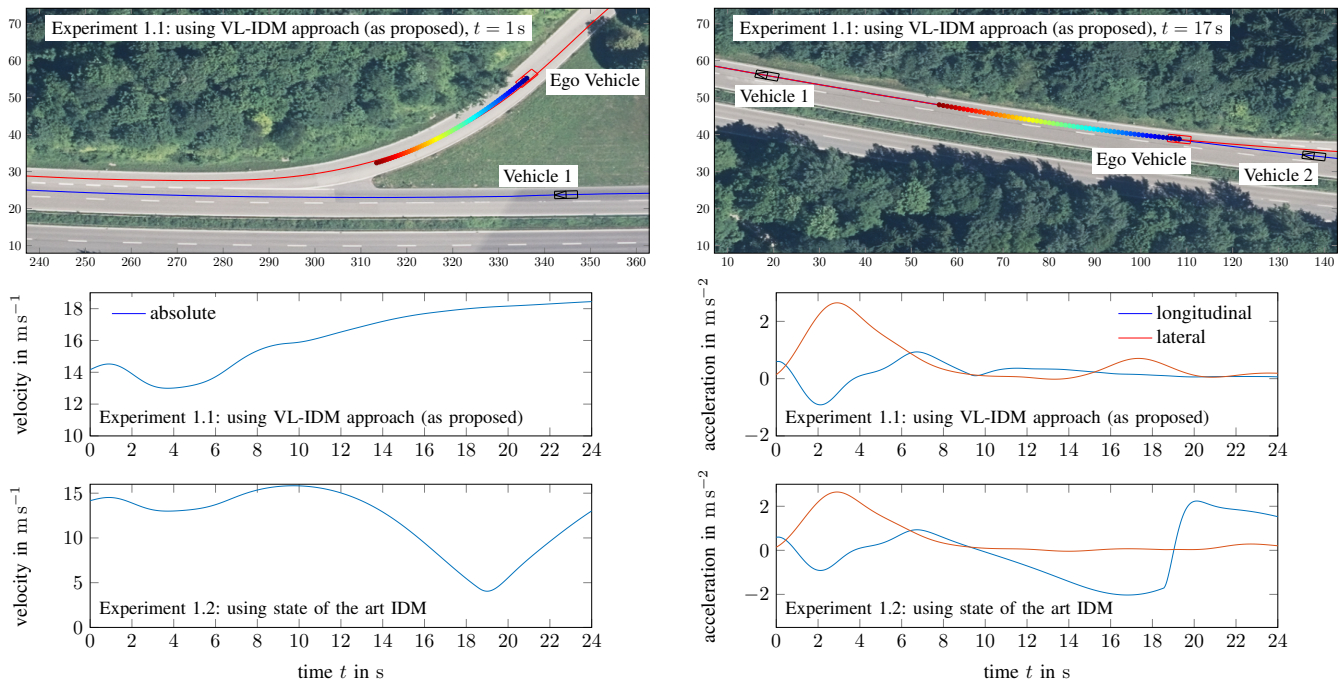
%
		\begin{subfigure}[t]{0.5\textwidth}
			\input{brOverview1Wide.tex}
		\end{subfigure}
		\begin{subfigure}[t]{0.5\textwidth}
			\input{brOverview3Wide.tex}
		\end{subfigure}
		\begin{subfigure}[t]{0.5\textwidth}
%
%
\definecolor{mycolor1}{rgb}{0.00000,0.44700,0.74100}%
\begin{tikzpicture}[scale=0.75]

\begin{axis}[%
name=MyAxis,
width=3.5in,
height=1in,
at={(0.758in,0.481in)},
scale only axis,
xmin=0,
xmax=24,
ymin=10,
ymax=19,
axis background/.style={fill=white}, 
ylabel={velocity in $\SI{}{\meter\per\second}$},
legend cell align=left,
legend pos= north west,
legend style={draw=none},
legend entries={absolute}
]

\addlegendimage{no markers,blue}

\addplot [color=mycolor1, forget plot]
  table[row sep=crcr]{%
0	14.1646\\
0.05	14.1946\\
0.1	14.2246\\
0.15	14.2545\\
0.2	14.2839\\
0.25	14.3126\\
0.3	14.3403\\
0.35	14.3668\\
0.4	14.3917\\
0.45	14.415\\
0.5	14.4364\\
0.55	14.4558\\
0.6	14.4729\\
0.65	14.4876\\
0.7	14.4998\\
0.75	14.5094\\
0.8	14.5162\\
0.85	14.5202\\
0.9	14.5214\\
0.95	14.5197\\
1	14.5151\\
1.05	14.5075\\
1.1	14.497\\
1.15	14.4836\\
1.2	14.4674\\
1.25	14.4483\\
1.3	14.4265\\
1.35	14.4021\\
1.4	14.3752\\
1.45	14.3458\\
1.5	14.3142\\
1.55	14.2803\\
1.6	14.2445\\
1.65	14.2068\\
1.7	14.1674\\
1.75	14.1265\\
1.8	14.0843\\
1.85	14.041\\
1.9	13.9966\\
1.95	13.9516\\
2	13.9059\\
2.05	13.8599\\
2.1	13.8137\\
2.15	13.7676\\
2.2	13.7216\\
2.25	13.6761\\
2.3	13.6311\\
2.35	13.5869\\
2.4	13.5436\\
2.45	13.5014\\
2.5	13.4604\\
2.55	13.4208\\
2.6	13.3826\\
2.65	13.3461\\
2.7	13.3112\\
2.75	13.2781\\
2.8	13.2468\\
2.85	13.2174\\
2.9	13.1899\\
2.95	13.1644\\
3	13.1408\\
3.05	13.1192\\
3.1	13.0995\\
3.15	13.0818\\
3.2	13.0659\\
3.25	13.0518\\
3.3	13.0395\\
3.35	13.0289\\
3.4	13.02\\
3.45	13.0126\\
3.5	13.0066\\
3.55	13.0021\\
3.6	12.9988\\
3.65	12.9968\\
3.7	12.996\\
3.75	12.9961\\
3.8	12.9973\\
3.85	12.9994\\
3.9	13.0023\\
3.95	13.0059\\
4	13.0103\\
4.05	13.0153\\
4.1	13.021\\
4.15	13.0272\\
4.2	13.0339\\
4.25	13.0411\\
4.3	13.0488\\
4.35	13.057\\
4.4	13.0656\\
4.45	13.0746\\
4.5	13.0841\\
4.55	13.0941\\
4.6	13.1045\\
4.65	13.1154\\
4.7	13.1268\\
4.75	13.1387\\
4.8	13.1512\\
4.85	13.1643\\
4.9	13.178\\
4.95	13.1924\\
5	13.2075\\
5.05	13.2233\\
5.1	13.24\\
5.15	13.2574\\
5.2	13.2758\\
5.25	13.295\\
5.3	13.3153\\
5.35	13.3365\\
5.4	13.3588\\
5.45	13.3822\\
5.5	13.4068\\
5.55	13.4324\\
5.6	13.4593\\
5.65	13.4874\\
5.7	13.5167\\
5.75	13.5472\\
5.8	13.579\\
5.85	13.612\\
5.9	13.6463\\
5.95	13.6818\\
6	13.7184\\
6.05	13.7563\\
6.1	13.7952\\
6.15	13.8353\\
6.2	13.8763\\
6.25	13.9184\\
6.3	13.9612\\
6.35	14.005\\
6.4	14.0494\\
6.45	14.0944\\
6.5	14.1399\\
6.55	14.1859\\
6.6	14.2322\\
6.65	14.2787\\
6.7	14.3253\\
6.75	14.3719\\
6.8	14.4184\\
6.85	14.4648\\
6.9	14.5109\\
6.95	14.5567\\
7	14.602\\
7.05	14.6469\\
7.1	14.6912\\
7.15	14.7349\\
7.2	14.778\\
7.25	14.8204\\
7.3	14.862\\
7.35	14.9028\\
7.4	14.9428\\
7.45	14.982\\
7.5	15.0203\\
7.55	15.0577\\
7.6	15.0942\\
7.65	15.1298\\
7.7	15.1645\\
7.75	15.1983\\
7.8	15.2311\\
7.85	15.263\\
7.9	15.2939\\
7.95	15.3239\\
8	15.353\\
8.05	15.3812\\
8.1	15.4084\\
8.15	15.4347\\
8.2	15.4601\\
8.25	15.4846\\
8.3	15.5082\\
8.35	15.5309\\
8.4	15.5527\\
8.45	15.5737\\
8.5	15.5938\\
8.55	15.613\\
8.6	15.6314\\
8.65	15.6489\\
8.7	15.6656\\
8.75	15.6814\\
8.8	15.6964\\
8.85	15.7106\\
8.9	15.724\\
8.95	15.7366\\
9	15.7484\\
9.05	15.7593\\
9.1	15.7695\\
9.15	15.7789\\
9.2	15.7874\\
9.25	15.7952\\
9.3	15.8022\\
9.35	15.8085\\
9.4	15.8141\\
9.45	15.8195\\
9.5	15.8246\\
9.55	15.8298\\
9.6	15.835\\
9.65	15.8405\\
9.7	15.8463\\
9.75	15.8524\\
9.8	15.8591\\
9.85	15.8663\\
9.9	15.874\\
9.95	15.8823\\
10	15.8912\\
10.05	15.9008\\
10.1	15.9109\\
10.15	15.9215\\
10.2	15.9328\\
10.25	15.9446\\
10.3	15.9569\\
10.35	15.9698\\
10.4	15.9832\\
10.45	15.9971\\
10.5	16.0114\\
10.55	16.0262\\
10.6	16.0413\\
10.65	16.0568\\
10.7	16.0726\\
10.75	16.0887\\
10.8	16.105\\
10.85	16.1217\\
10.9	16.1386\\
10.95	16.1556\\
11	16.1729\\
11.05	16.1904\\
11.1	16.208\\
11.15	16.2257\\
11.2	16.2436\\
11.25	16.2616\\
11.3	16.2796\\
11.35	16.2978\\
11.4	16.3159\\
11.45	16.3341\\
11.5	16.3523\\
11.55	16.3705\\
11.6	16.3887\\
11.65	16.4069\\
11.7	16.425\\
11.75	16.4431\\
11.8	16.461\\
11.85	16.4789\\
11.9	16.4967\\
11.95	16.5145\\
12	16.5321\\
12.05	16.5496\\
12.1	16.567\\
12.15	16.5843\\
12.2	16.6016\\
12.25	16.6188\\
12.3	16.6359\\
12.35	16.653\\
12.4	16.6701\\
12.45	16.6871\\
12.5	16.7041\\
12.55	16.7211\\
12.6	16.7381\\
12.65	16.755\\
12.7	16.772\\
12.75	16.7889\\
12.8	16.8058\\
12.85	16.8227\\
12.9	16.8396\\
12.95	16.8565\\
13	16.8733\\
13.05	16.89\\
13.1	16.9068\\
13.15	16.9235\\
13.2	16.9402\\
13.25	16.9569\\
13.3	16.9735\\
13.35	16.9901\\
13.4	17.0066\\
13.45	17.0231\\
13.5	17.0395\\
13.55	17.0559\\
13.6	17.0721\\
13.65	17.0883\\
13.7	17.1044\\
13.75	17.1204\\
13.8	17.1363\\
13.85	17.1521\\
13.9	17.1677\\
13.95	17.1832\\
14	17.1985\\
14.05	17.2137\\
14.1	17.2288\\
14.15	17.2437\\
14.2	17.2584\\
14.25	17.273\\
14.3	17.2874\\
14.35	17.3017\\
14.4	17.3159\\
14.45	17.3299\\
14.5	17.3437\\
14.55	17.3574\\
14.6	17.3709\\
14.65	17.3842\\
14.7	17.3974\\
14.75	17.4104\\
14.8	17.4233\\
14.85	17.436\\
14.9	17.4486\\
14.95	17.461\\
15	17.4733\\
15.05	17.4854\\
15.1	17.4973\\
15.15	17.5092\\
15.2	17.5208\\
15.25	17.5324\\
15.3	17.5438\\
15.35	17.555\\
15.4	17.5661\\
15.45	17.577\\
15.5	17.5878\\
15.55	17.5984\\
15.6	17.6088\\
15.65	17.6191\\
15.7	17.6292\\
15.75	17.6392\\
15.8	17.6489\\
15.85	17.6585\\
15.9	17.6679\\
15.95	17.6772\\
16	17.6862\\
16.05	17.6952\\
16.1	17.7039\\
16.15	17.7126\\
16.2	17.7211\\
16.25	17.7295\\
16.3	17.7378\\
16.35	17.746\\
16.4	17.7541\\
16.45	17.7622\\
16.5	17.7702\\
16.55	17.7781\\
16.6	17.786\\
16.65	17.7939\\
16.7	17.8017\\
16.75	17.8095\\
16.8	17.8172\\
16.85	17.8249\\
16.9	17.8326\\
16.95	17.8403\\
17	17.8479\\
17.05	17.8554\\
17.1	17.8629\\
17.15	17.8704\\
17.2	17.8778\\
17.25	17.8851\\
17.3	17.8924\\
17.35	17.8996\\
17.4	17.9068\\
17.45	17.9139\\
17.5	17.9209\\
17.55	17.9278\\
17.6	17.9346\\
17.65	17.9414\\
17.7	17.948\\
17.75	17.9546\\
17.8	17.9611\\
17.85	17.9675\\
17.9	17.9738\\
17.95	17.98\\
18	17.9862\\
18.05	17.9922\\
18.1	17.9981\\
18.15	18.004\\
18.2	18.0098\\
18.25	18.0155\\
18.3	18.0211\\
18.35	18.0267\\
18.4	18.0322\\
18.45	18.0376\\
18.5	18.0429\\
18.55	18.0482\\
18.6	18.0534\\
18.65	18.0585\\
18.7	18.0634\\
18.75	18.0683\\
18.8	18.0731\\
18.85	18.0778\\
18.9	18.0824\\
18.95	18.0869\\
19	18.0912\\
19.05	18.0955\\
19.1	18.0996\\
19.15	18.1037\\
19.2	18.1076\\
19.25	18.1114\\
19.3	18.1152\\
19.35	18.1188\\
19.4	18.1223\\
19.45	18.1258\\
19.5	18.1292\\
19.55	18.1325\\
19.6	18.1358\\
19.65	18.139\\
19.7	18.1421\\
19.75	18.1452\\
19.8	18.1483\\
19.85	18.1513\\
19.9	18.1543\\
19.95	18.1573\\
20	18.1603\\
20.05	18.1633\\
20.1	18.1663\\
20.15	18.1694\\
20.2	18.1724\\
20.25	18.1755\\
20.3	18.1786\\
20.35	18.1817\\
20.4	18.1849\\
20.45	18.1881\\
20.5	18.1914\\
20.55	18.1947\\
20.6	18.198\\
20.65	18.2013\\
20.7	18.2047\\
20.75	18.2081\\
20.8	18.2115\\
20.85	18.2149\\
20.9	18.2184\\
20.95	18.222\\
21	18.2255\\
21.05	18.2291\\
21.1	18.2327\\
21.15	18.2363\\
21.2	18.2399\\
21.25	18.2435\\
21.3	18.2472\\
21.35	18.2508\\
21.4	18.2545\\
21.45	18.2582\\
21.5	18.2619\\
21.55	18.2656\\
21.6	18.2693\\
21.65	18.273\\
21.7	18.2767\\
21.75	18.2804\\
21.8	18.2842\\
21.85	18.2879\\
21.9	18.2916\\
21.95	18.2953\\
22	18.299\\
22.05	18.3027\\
22.1	18.3065\\
22.15	18.3102\\
22.2	18.3139\\
22.25	18.3177\\
22.3	18.3214\\
22.35	18.3252\\
22.4	18.3289\\
22.45	18.3326\\
22.5	18.3364\\
22.55	18.3401\\
22.6	18.3439\\
22.65	18.3476\\
22.7	18.3514\\
22.75	18.3551\\
22.8	18.3589\\
22.85	18.3626\\
22.9	18.3663\\
22.95	18.37\\
23	18.3738\\
23.05	18.3775\\
23.1	18.3812\\
23.15	18.3849\\
23.2	18.3886\\
23.25	18.3923\\
23.3	18.3959\\
23.35	18.3996\\
23.4	18.4032\\
23.45	18.4068\\
23.5	18.4104\\
23.55	18.4139\\
23.6	18.4175\\
23.65	18.421\\
23.7	18.4244\\
23.75	18.4278\\
23.8	18.4312\\
23.85	18.4345\\
23.9	18.4377\\
23.95	18.4409\\
24	18.444\\
};
\end{axis}
\node[above right, align=center, text=black]
at (MyAxis.south west) {%
	\scriptsize{Experiment 1.1: using VL-IDM approach (as proposed)}};
\end{tikzpicture}%
		\end{subfigure}
		\begin{subfigure}[t]{0.5\textwidth}
			\input{tikzPlots/br_lat_lon_acc}
		\end{subfigure}
		\begin{subfigure}[t]{0.5\textwidth}
%
%
\definecolor{mycolor1}{rgb}{0.00000,0.44700,0.74100}%
\begin{tikzpicture}[scale=0.75]

\begin{axis}[%
width=3.5in,
height=1in,
at={(0.758in,0.481in)},
scale only axis,
xmin=0,
xmax=24,
ymin=0,
ymax=16,
axis background/.style={fill=white},
ylabel={velocity in $\SI{}{\meter\per\second}$},
xlabel={time $t$ in $\SI{}{\second}$}
]
\addplot [color=mycolor1, forget plot]
  table[row sep=crcr]{%
0	14.1646\\
0.05	14.1946\\
0.1	14.2246\\
0.15	14.2545\\
0.2	14.2839\\
0.25	14.3126\\
0.3	14.3403\\
0.35	14.3668\\
0.4	14.3917\\
0.45	14.415\\
0.5	14.4364\\
0.55	14.4558\\
0.6	14.4729\\
0.65	14.4876\\
0.7	14.4998\\
0.75	14.5094\\
0.8	14.5162\\
0.85	14.5202\\
0.9	14.5214\\
0.95	14.5197\\
1	14.5151\\
1.05	14.5075\\
1.1	14.497\\
1.15	14.4836\\
1.2	14.4674\\
1.25	14.4483\\
1.3	14.4265\\
1.35	14.4021\\
1.4	14.3752\\
1.45	14.3458\\
1.5	14.3142\\
1.55	14.2803\\
1.6	14.2445\\
1.65	14.2068\\
1.7	14.1674\\
1.75	14.1265\\
1.8	14.0843\\
1.85	14.041\\
1.9	13.9966\\
1.95	13.9516\\
2	13.9059\\
2.05	13.8599\\
2.1	13.8137\\
2.15	13.7676\\
2.2	13.7216\\
2.25	13.6761\\
2.3	13.6311\\
2.35	13.5869\\
2.4	13.5436\\
2.45	13.5014\\
2.5	13.4604\\
2.55	13.4208\\
2.6	13.3826\\
2.65	13.3461\\
2.7	13.3112\\
2.75	13.2781\\
2.8	13.2468\\
2.85	13.2174\\
2.9	13.1899\\
2.95	13.1644\\
3	13.1408\\
3.05	13.1192\\
3.1	13.0995\\
3.15	13.0818\\
3.2	13.0659\\
3.25	13.0518\\
3.3	13.0395\\
3.35	13.0289\\
3.4	13.02\\
3.45	13.0126\\
3.5	13.0066\\
3.55	13.0021\\
3.6	12.9988\\
3.65	12.9968\\
3.7	12.996\\
3.75	12.9961\\
3.8	12.9973\\
3.85	12.9994\\
3.9	13.0023\\
3.95	13.0059\\
4	13.0103\\
4.05	13.0153\\
4.1	13.021\\
4.15	13.0272\\
4.2	13.0339\\
4.25	13.0411\\
4.3	13.0488\\
4.35	13.057\\
4.4	13.0656\\
4.45	13.0746\\
4.5	13.0841\\
4.55	13.0941\\
4.6	13.1045\\
4.65	13.1154\\
4.7	13.1268\\
4.75	13.1387\\
4.8	13.1512\\
4.85	13.1643\\
4.9	13.178\\
4.95	13.1924\\
5	13.2075\\
5.05	13.2233\\
5.1	13.24\\
5.15	13.2574\\
5.2	13.2758\\
5.25	13.295\\
5.3	13.3153\\
5.35	13.3365\\
5.4	13.3588\\
5.45	13.3822\\
5.5	13.4068\\
5.55	13.4324\\
5.6	13.4593\\
5.65	13.4874\\
5.7	13.5167\\
5.75	13.5472\\
5.8	13.579\\
5.85	13.612\\
5.9	13.6463\\
5.95	13.6818\\
6	13.7184\\
6.05	13.7563\\
6.1	13.7952\\
6.15	13.8353\\
6.2	13.8763\\
6.25	13.9184\\
6.3	13.9612\\
6.35	14.005\\
6.4	14.0494\\
6.45	14.0944\\
6.5	14.1399\\
6.55	14.1859\\
6.6	14.2322\\
6.65	14.2787\\
6.7	14.3253\\
6.75	14.3719\\
6.8	14.4184\\
6.85	14.4648\\
6.9	14.5109\\
6.95	14.5567\\
7	14.602\\
7.05	14.6469\\
7.1	14.6912\\
7.15	14.7349\\
7.2	14.778\\
7.25	14.8204\\
7.3	14.862\\
7.35	14.9028\\
7.4	14.9428\\
7.45	14.982\\
7.5	15.0203\\
7.55	15.0577\\
7.6	15.0942\\
7.65	15.1298\\
7.7	15.1645\\
7.75	15.1983\\
7.8	15.2311\\
7.85	15.263\\
7.9	15.2939\\
7.95	15.3239\\
8	15.353\\
8.05	15.3812\\
8.1	15.4084\\
8.15	15.4347\\
8.2	15.4601\\
8.25	15.4846\\
8.3	15.5082\\
8.35	15.5309\\
8.4	15.5527\\
8.45	15.5737\\
8.5	15.5938\\
8.55	15.613\\
8.6	15.6314\\
8.65	15.6489\\
8.7	15.6656\\
8.75	15.6814\\
8.8	15.6964\\
8.85	15.7106\\
8.9	15.724\\
8.95	15.7366\\
9	15.7484\\
9.05	15.7593\\
9.1	15.7695\\
9.15	15.7789\\
9.2	15.7874\\
9.25	15.7952\\
9.3	15.8022\\
9.35	15.8085\\
9.4	15.8139\\
9.45	15.8186\\
9.5	15.8225\\
9.55	15.8256\\
9.6	15.828\\
9.65	15.8295\\
9.7	15.8304\\
9.75	15.8304\\
9.8	15.8297\\
9.85	15.8282\\
9.9	15.8259\\
9.95	15.8229\\
10	15.8191\\
10.05	15.8146\\
10.1	15.8092\\
10.15	15.8031\\
10.2	15.7963\\
10.25	15.7887\\
10.3	15.7803\\
10.35	15.7711\\
10.4	15.7612\\
10.45	15.7505\\
10.5	15.739\\
10.55	15.7268\\
10.6	15.7138\\
10.65	15.7\\
10.7	15.6854\\
10.75	15.6701\\
10.8	15.654\\
10.85	15.6371\\
10.9	15.6194\\
10.95	15.601\\
11	15.5817\\
11.05	15.5617\\
11.1	15.5409\\
11.15	15.5193\\
11.2	15.497\\
11.25	15.4738\\
11.3	15.4498\\
11.35	15.4251\\
11.4	15.3995\\
11.45	15.3731\\
11.5	15.346\\
11.55	15.318\\
11.6	15.2892\\
11.65	15.2596\\
11.7	15.2292\\
11.75	15.198\\
11.8	15.166\\
11.85	15.1331\\
11.9	15.0994\\
11.95	15.0649\\
12	15.0295\\
12.05	14.9933\\
12.1	14.9563\\
12.15	14.9184\\
12.2	14.8797\\
12.25	14.8402\\
12.3	14.7998\\
12.35	14.7585\\
12.4	14.7164\\
12.45	14.6734\\
12.5	14.6295\\
12.55	14.5848\\
12.6	14.5392\\
12.65	14.4928\\
12.7	14.4454\\
12.75	14.3972\\
12.8	14.3481\\
12.85	14.2981\\
12.9	14.2473\\
12.95	14.1955\\
13	14.1429\\
13.05	14.0893\\
13.1	14.0349\\
13.15	13.9795\\
13.2	13.9233\\
13.25	13.8661\\
13.3	13.8081\\
13.35	13.7491\\
13.4	13.6893\\
13.45	13.6285\\
13.5	13.5668\\
13.55	13.5042\\
13.6	13.4407\\
13.65	13.3763\\
13.7	13.311\\
13.75	13.2448\\
13.8	13.1777\\
13.85	13.1097\\
13.9	13.0408\\
13.95	12.971\\
14	12.9003\\
14.05	12.8287\\
14.1	12.7562\\
14.15	12.6829\\
14.2	12.6086\\
14.25	12.5335\\
14.3	12.4576\\
14.35	12.3807\\
14.4	12.303\\
14.45	12.2245\\
14.5	12.1451\\
14.55	12.0649\\
14.6	11.9839\\
14.65	11.9021\\
14.7	11.8195\\
14.75	11.736\\
14.8	11.6518\\
14.85	11.5669\\
14.9	11.4812\\
14.95	11.3947\\
15	11.3075\\
15.05	11.2196\\
15.1	11.131\\
15.15	11.0418\\
15.2	10.9518\\
15.25	10.8612\\
15.3	10.77\\
15.35	10.6781\\
15.4	10.5857\\
15.45	10.4926\\
15.5	10.399\\
15.55	10.3048\\
15.6	10.2101\\
15.65	10.1149\\
15.7	10.0192\\
15.75	9.92298\\
15.8	9.82633\\
15.85	9.72922\\
15.9	9.6317\\
15.95	9.53379\\
16	9.43549\\
16.05	9.33682\\
16.1	9.23781\\
16.15	9.13849\\
16.2	9.03887\\
16.25	8.93897\\
16.3	8.83881\\
16.35	8.73842\\
16.4	8.63781\\
16.45	8.53702\\
16.5	8.43606\\
16.55	8.33496\\
16.6	8.23374\\
16.65	8.13243\\
16.7	8.03105\\
16.75	7.92964\\
16.8	7.8282\\
16.85	7.72678\\
16.9	7.62538\\
16.95	7.52405\\
17	7.4228\\
17.05	7.32166\\
17.1	7.22066\\
17.15	7.11982\\
17.2	7.01917\\
17.25	6.91873\\
17.3	6.81853\\
17.35	6.7186\\
17.4	6.61896\\
17.45	6.51964\\
17.5	6.42067\\
17.55	6.32207\\
17.6	6.22388\\
17.65	6.12611\\
17.7	6.02879\\
17.75	5.93197\\
17.8	5.83566\\
17.85	5.73988\\
17.9	5.64467\\
17.95	5.55006\\
18	5.45608\\
18.05	5.36274\\
18.1	5.27007\\
18.15	5.17811\\
18.2	5.08688\\
18.25	4.9964\\
18.3	4.9067\\
18.35	4.8178\\
18.4	4.72972\\
18.45	4.64249\\
18.5	4.55612\\
18.55	4.47065\\
18.6	4.38863\\
18.65	4.31226\\
18.7	4.24344\\
18.75	4.18377\\
18.8	4.13406\\
18.85	4.09498\\
18.9	4.06702\\
18.95	4.05054\\
19	4.0454\\
19.05	4.05136\\
19.1	4.06816\\
19.15	4.0955\\
19.2	4.13275\\
19.25	4.17927\\
19.3	4.23447\\
19.35	4.29775\\
19.4	4.36835\\
19.45	4.44557\\
19.5	4.52872\\
19.55	4.61718\\
19.6	4.71029\\
19.65	4.8074\\
19.7	4.90795\\
19.75	5.01139\\
19.8	5.11723\\
19.85	5.22501\\
19.9	5.33428\\
19.95	5.44467\\
20	5.55586\\
20.05	5.66754\\
20.1	5.77944\\
20.15	5.89132\\
20.2	6.003\\
20.25	6.11433\\
20.3	6.22517\\
20.35	6.33538\\
20.4	6.4449\\
20.45	6.55368\\
20.5	6.66166\\
20.55	6.76878\\
20.6	6.87506\\
20.65	6.9805\\
20.7	7.08509\\
20.75	7.18885\\
20.8	7.2918\\
20.85	7.39399\\
20.9	7.49544\\
20.95	7.59617\\
21	7.69623\\
21.05	7.79566\\
21.1	7.8945\\
21.15	7.99277\\
21.2	8.09052\\
21.25	8.18779\\
21.3	8.2846\\
21.35	8.38098\\
21.4	8.47696\\
21.45	8.57257\\
21.5	8.66783\\
21.55	8.76276\\
21.6	8.85738\\
21.65	8.9517\\
21.7	9.04574\\
21.75	9.1395\\
21.8	9.23299\\
21.85	9.32623\\
21.9	9.41921\\
21.95	9.51194\\
22	9.60441\\
22.05	9.69663\\
22.1	9.78859\\
22.15	9.88029\\
22.2	9.97172\\
22.25	10.0629\\
22.3	10.1538\\
22.35	10.2444\\
22.4	10.3347\\
22.45	10.4247\\
22.5	10.5144\\
22.55	10.6037\\
22.6	10.6928\\
22.65	10.7815\\
22.7	10.8699\\
22.75	10.9579\\
22.8	11.0456\\
22.85	11.1329\\
22.9	11.2198\\
22.95	11.3064\\
23	11.3925\\
23.05	11.4782\\
23.1	11.5635\\
23.15	11.6484\\
23.2	11.7328\\
23.25	11.8168\\
23.3	11.9004\\
23.35	11.9835\\
23.4	12.0661\\
23.45	12.1482\\
23.5	12.2299\\
23.55	12.311\\
23.6	12.3917\\
23.65	12.4718\\
23.7	12.5514\\
23.75	12.6305\\
23.8	12.7091\\
23.85	12.7871\\
23.9	12.8646\\
23.95	12.9415\\
24	13.0179\\
};
\end{axis}
\node[above right, align=center, text=black]
at (MyAxis.south west) {%
	\scriptsize{Experiment 1.2: using state of the art IDM}};
\end{tikzpicture}%
		\end{subfigure}
		\begin{subfigure}[t]{0.5\textwidth}
			\input{tikzPlots/br_lat_lon_acc_no_virtual}
		\end{subfigure}
		\caption{Merging Scenario, where two experiments are regarded. In Experiment 1.1, the presented approach is applied using the parameters shown in Table \ref{tab:parameters}. 
			The first row depicts the corresponding scene true to scale in a local map aligned to Universal Transverse Mercator coordinates.
			The ego vehicle and ego route are indicated by red color. The states of the planned ego trajectory at the presented time step are shown by dots.
			Beginning in black with the state for $t$ and ending with the state planned for $t+(N_\text{opt}-1) \Delta t$ in red. The second row shows the velocity and accelerations of the ego vehicle for Experiment 1.1. In Experiment 1.2, the same scenario is regarded, however, no reference trajectory generated with the VL approach is utilized. The third row depicts the corresponding velocity and acceleration. It can be seen that the ego vehicle is not able to merge between the vehicles and, therefore, has to drastically reduce the velocity. As a result, it is emphasized that the VL approach ensures much better merge behavior and, consequently, outperforms the state of the art.}
		\label{fig:brOverview}
	\end{figure*}

	The proposed framework is evaluated on a Intel Xeon E5-1660 CPU with 3.2 GHz and implemented in C++.
	The sequential quadratic programming-method of the WORHP solver \cite{Bueskens2013} is utilized to solve the optimization problem shown in Section \ref{subsec:optimization}. 
	In order to demonstrate the versatility of the trajectory planning framework, we consider two different intersection scenarios, where both scenarios are regarded twice. The real-world center lines are extracted using the high precision Digital Map of Ulm University \cite{Kunz2015}. Further, Table \ref{tab:parameters} showcases the parametrization of variables. The ego trajectory is replanned every $\Delta t_\text{replan} = \SI{0.2}{\second}$. The other vehicles are simulated using a double integrator where the system input is the acceleration that is uniformly distributed over $\SI{1}{\meter\per\second\squared}$ and $\SI{-1}{\meter\per\second\squared}$ in each time step. Further, the initial speed may also differ from the expected target speed.
	
	{\renewcommand{\arraystretch}{1.0} 
		\begin{table}
			\caption{Parameters used for evaluation}
			\label{tab:parameters}
			{\sisetup{per-mode=fraction}
				\begin{center}
					\resizebox{\columnwidth}{!}{%
						\begin{tabular}{c c c c c c c c}
							\hline
							$ a_\text{IDM}$ &$\SI{2}{\meter\per\second\squared}$ &	&$w_c$ & 1 &&$N_\text{ref}$&201\\
							$b$ & $\SI{4}{\meter\per\second\squared}$ &				&$w_p$& 1 &&$N_\text{opt}$&60\\
							$s_0$ & $\SI{4}{\meter}$ & &$w_{\text{spatial}}$& 1 &	&$\delta_r$& $0.1$\\
							$T$ & $\SI{2.5}{\second}$ &				&$w_{\text{acc}}$&0.1 &					&$H_a^\text{const}$&$\SI{0.25}{\meter\per\second\squared}$\\
							$\delta $ & 4 & &$w_{\text{jerk}}$ &0.1 &&$H_\text{ttc}^\text{const}$&$\SI{0.25}{\second}$\\
							$\Delta t$& $\SI{0.05}{\second}$&&$w_{\text{snap}}$ &0.1 &&$\Delta 
							a_\text{max}$&$\SI{0.9}{\meter\per\second\squared}$\\
							$\Delta t_\text{replan}$&$\SI{0.2}{\second}$&&$a_\text{max}$&$\SI{5}{\meter\per\second\squared}$&&$\Delta t_\text{max}$&$\SI{1}{\second}$\\
							\hline
						\end{tabular}
					}
				\end{center}
			}
			\vspace{-6mm}
		\end{table}
	}
	
	The first intersection represents a merging scenario on rural roads where the speed limit or respectively the target speed is $\SI{70}{\kilo\meter\per\hour}$, cf. \figurename~\ref{fig:brOverview}.
	The ego vehicle is approaching from the top following a curve and merges into another lane. This lane is occupied by Vehicle 1 and Vehicle 2 with an initial speed of $\SI{70}{\kilo\meter\per\hour}$. In Experiment 1.1, the proposed method is used. Thereby, the ego vehicle first adapts the speed appropriate to the curvature of the lane in order to restrict the lateral acceleration. Around $\SI{7}{\second}$, a reduction of the longitudinal acceleration is performed to be able to smoothly merge behind Vehicle 1. Then the longitudinal acceleration is slightly increased to catch up with the speed of the other vehicles. In the end, a safe state between the other vehicles is reached.
	Hence, this demonstrates a smooth merging maneuver with a suitable distance to the leading vehicle (Vehicle 1) and appropriate costs for Vehicle 2.
	To emphasize the advantages of the reference trajectory generated with the VL approach shown in Section \ref{subsec:referenceTraj}, the same scenario is regarded again in Experiment 1.2. 
	However, only reference trajectories, or respectively behavior options, are used without using a VL. \figurename~\ref{fig:brOverview} presents the corresponding velocity and accelerations for both experiments.
	It can bee seen that without the VL approach, the ego vehicle is not able to merge behind Vehicle 1 and, therefore, the velocity is drastically reduced.
	Once Vehicle 2 has passed, the acceleration is strongly increased. This example clearly indicates that the presented reference trajectory generation with the VL approach enables a much better merge behavior and, as a result, outperforms the state of the art.
	
\begin{figure*}[t]
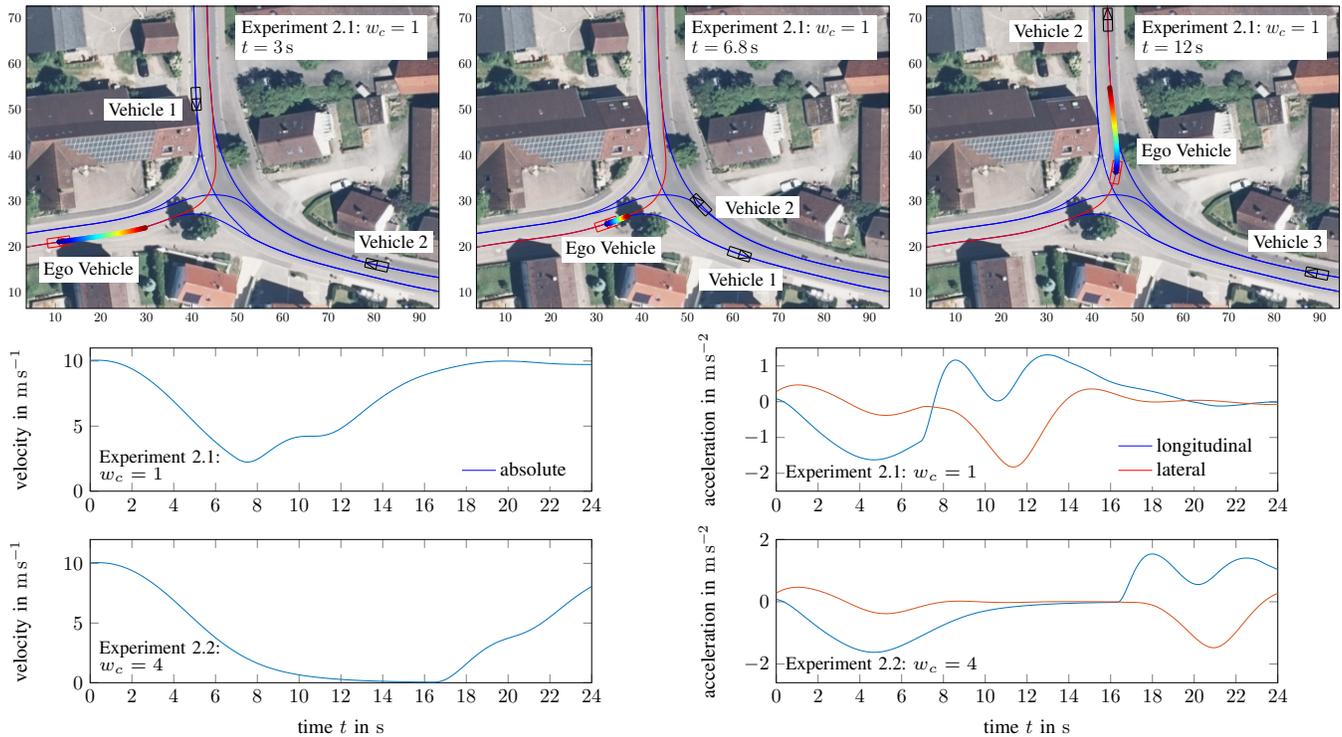
%
		\begin{subfigure}[t]{0.33\textwidth}
			\input{lehrOverview1.tex}
		\end{subfigure}
		\begin{subfigure}[t]{0.33\textwidth}
			\input{lehrOverview2.tex}
		\end{subfigure}
		\begin{subfigure}[t]{0.33\textwidth}
			\input{lehrOverview3.tex}
		\end{subfigure}
		\begin{subfigure}[t]{0.5\textwidth}
%
%
\definecolor{mycolor1}{rgb}{0.00000,0.44700,0.74100}%
\begin{tikzpicture}[scale=0.75]

\begin{axis}[%
width=3.5in,
height=1in,
at={(0.758in,0.481in)},
scale only axis,
xmin=0,
xmax=24,
ymin=0,
ymax=11,
axis background/.style={fill=white},
ylabel={velocity in $\SI{}{\meter\per\second}$},
legend cell align=left,
legend pos= south east,
legend style={draw=none},
legend entries={absolute}
]

\addlegendimage{no markers,blue}

\addplot [color=mycolor1, forget plot]
  table[row sep=crcr]{%
0	10.0482\\
0.05	10.0518\\
0.1	10.0549\\
0.15	10.0575\\
0.2	10.0596\\
0.25	10.061\\
0.3	10.0617\\
0.35	10.0616\\
0.4	10.0607\\
0.45	10.0588\\
0.5	10.056\\
0.55	10.0522\\
0.6	10.0473\\
0.65	10.0413\\
0.7	10.0341\\
0.75	10.0258\\
0.8	10.0163\\
0.85	10.0055\\
0.9	9.99346\\
0.95	9.98015\\
1	9.96554\\
1.05	9.94961\\
1.1	9.93236\\
1.15	9.91378\\
1.2	9.89386\\
1.25	9.8726\\
1.3	9.84998\\
1.35	9.82603\\
1.4	9.80074\\
1.45	9.7741\\
1.5	9.74612\\
1.55	9.71681\\
1.6	9.68618\\
1.65	9.65423\\
1.7	9.62096\\
1.75	9.58639\\
1.8	9.55053\\
1.85	9.51337\\
1.9	9.47494\\
1.95	9.43526\\
2	9.39431\\
2.05	9.35212\\
2.1	9.3087\\
2.15	9.26406\\
2.2	9.21822\\
2.25	9.17118\\
2.3	9.12295\\
2.35	9.07357\\
2.4	9.02304\\
2.45	8.97136\\
2.5	8.91856\\
2.55	8.86466\\
2.6	8.80966\\
2.65	8.75358\\
2.7	8.69644\\
2.75	8.63826\\
2.8	8.57904\\
2.85	8.51881\\
2.9	8.45758\\
2.95	8.39538\\
3	8.33221\\
3.05	8.26809\\
3.1	8.20304\\
3.15	8.13709\\
3.2	8.07025\\
3.25	8.00253\\
3.3	7.93397\\
3.35	7.86458\\
3.4	7.79439\\
3.45	7.72341\\
3.5	7.65167\\
3.55	7.5792\\
3.6	7.50602\\
3.65	7.43214\\
3.7	7.35761\\
3.75	7.28245\\
3.8	7.20668\\
3.85	7.13033\\
3.9	7.05344\\
3.95	6.97604\\
4	6.89814\\
4.05	6.81979\\
4.1	6.74101\\
4.15	6.66185\\
4.2	6.58233\\
4.25	6.50248\\
4.3	6.42234\\
4.35	6.34195\\
4.4	6.26133\\
4.45	6.18053\\
4.5	6.09957\\
4.55	6.01851\\
4.6	5.93736\\
4.65	5.85617\\
4.7	5.77497\\
4.75	5.6938\\
4.8	5.61269\\
4.85	5.53169\\
4.9	5.45082\\
4.95	5.37012\\
5	5.28963\\
5.05	5.20937\\
5.1	5.12939\\
5.15	5.04972\\
5.2	4.97038\\
5.25	4.89141\\
5.3	4.81285\\
5.35	4.73471\\
5.4	4.65703\\
5.45	4.57984\\
5.5	4.50316\\
5.55	4.42702\\
5.6	4.35145\\
5.65	4.27646\\
5.7	4.20208\\
5.75	4.12833\\
5.8	4.05524\\
5.85	3.98281\\
5.9	3.91108\\
5.95	3.84005\\
6	3.76974\\
6.05	3.70017\\
6.1	3.63135\\
6.15	3.56329\\
6.2	3.49601\\
6.25	3.42951\\
6.3	3.36381\\
6.35	3.29892\\
6.4	3.23484\\
6.45	3.17158\\
6.5	3.10915\\
6.55	3.04755\\
6.6	2.98679\\
6.65	2.92686\\
6.7	2.86779\\
6.75	2.80956\\
6.8	2.75219\\
6.85	2.69566\\
6.9	2.63999\\
6.95	2.58518\\
7	2.53223\\
7.05	2.48203\\
7.1	2.43532\\
7.15	2.39274\\
7.2	2.35474\\
7.25	2.32172\\
7.3	2.29396\\
7.35	2.27168\\
7.4	2.255\\
7.45	2.24397\\
7.5	2.23858\\
7.55	2.23882\\
7.6	2.2445\\
7.65	2.25545\\
7.7	2.27148\\
7.75	2.29236\\
7.8	2.31781\\
7.85	2.34753\\
7.9	2.38125\\
7.95	2.41867\\
8	2.45946\\
8.05	2.5033\\
8.1	2.54986\\
8.15	2.59887\\
8.2	2.64999\\
8.25	2.70292\\
8.3	2.75737\\
8.35	2.81309\\
8.4	2.86978\\
8.45	2.92718\\
8.5	2.98503\\
8.55	3.04313\\
8.6	3.10122\\
8.65	3.15909\\
8.7	3.21656\\
8.75	3.27344\\
8.8	3.32956\\
8.85	3.38476\\
8.9	3.43889\\
8.95	3.49183\\
9	3.54345\\
9.05	3.59361\\
9.1	3.64222\\
9.15	3.68919\\
9.2	3.7344\\
9.25	3.77773\\
9.3	3.81911\\
9.35	3.85848\\
9.4	3.89576\\
9.45	3.93091\\
9.5	3.9639\\
9.55	3.99472\\
9.6	4.02334\\
9.65	4.04978\\
9.7	4.07405\\
9.75	4.0962\\
9.8	4.11625\\
9.85	4.13424\\
9.9	4.15025\\
9.95	4.16434\\
10	4.17661\\
10.05	4.18713\\
10.1	4.19602\\
10.15	4.20339\\
10.2	4.20938\\
10.25	4.21411\\
10.3	4.21772\\
10.35	4.22038\\
10.4	4.22224\\
10.45	4.22347\\
10.5	4.22424\\
10.55	4.22472\\
10.6	4.22511\\
10.65	4.22562\\
10.7	4.22643\\
10.75	4.22774\\
10.8	4.22974\\
10.85	4.23263\\
10.9	4.2366\\
10.95	4.24182\\
11	4.24847\\
11.05	4.25671\\
11.1	4.2667\\
11.15	4.27856\\
11.2	4.29242\\
11.25	4.30841\\
11.3	4.32661\\
11.35	4.34708\\
11.4	4.3699\\
11.45	4.3951\\
11.5	4.42271\\
11.55	4.45272\\
11.6	4.48511\\
11.65	4.51986\\
11.7	4.5569\\
11.75	4.59617\\
11.8	4.6376\\
11.85	4.68111\\
11.9	4.72661\\
11.95	4.77399\\
12	4.82317\\
12.05	4.87405\\
12.1	4.92652\\
12.15	4.98049\\
12.2	5.03585\\
12.25	5.09251\\
12.3	5.15037\\
12.35	5.20932\\
12.4	5.26928\\
12.45	5.33016\\
12.5	5.39187\\
12.55	5.45432\\
12.6	5.51743\\
12.65	5.5811\\
12.7	5.64525\\
12.75	5.70979\\
12.8	5.77463\\
12.85	5.83971\\
12.9	5.90496\\
12.95	5.97029\\
13	6.03564\\
13.05	6.10095\\
13.1	6.16615\\
13.15	6.23118\\
13.2	6.29597\\
13.25	6.36047\\
13.3	6.42461\\
13.35	6.48835\\
13.4	6.55161\\
13.45	6.61434\\
13.5	6.67649\\
13.55	6.73803\\
13.6	6.79891\\
13.65	6.85912\\
13.7	6.91864\\
13.75	6.97744\\
13.8	7.03554\\
13.85	7.09294\\
13.9	7.14965\\
13.95	7.20568\\
14	7.26104\\
14.05	7.31576\\
14.1	7.36985\\
14.15	7.42332\\
14.2	7.47618\\
14.25	7.52845\\
14.3	7.58012\\
14.35	7.6312\\
14.4	7.6817\\
14.45	7.73159\\
14.5	7.78088\\
14.55	7.82956\\
14.6	7.8776\\
14.65	7.925\\
14.7	7.97172\\
14.75	8.01776\\
14.8	8.06309\\
14.85	8.10771\\
14.9	8.15159\\
14.95	8.19475\\
15	8.23715\\
15.05	8.27879\\
15.1	8.31965\\
15.15	8.35974\\
15.2	8.39904\\
15.25	8.43753\\
15.3	8.47522\\
15.35	8.51209\\
15.4	8.54816\\
15.45	8.58343\\
15.5	8.61793\\
15.55	8.65167\\
15.6	8.68467\\
15.65	8.71695\\
15.7	8.74853\\
15.75	8.77944\\
15.8	8.8097\\
15.85	8.83932\\
15.9	8.86832\\
15.95	8.89673\\
16	8.92457\\
16.05	8.95186\\
16.1	8.97862\\
16.15	9.00488\\
16.2	9.03064\\
16.25	9.05594\\
16.3	9.08077\\
16.35	9.10517\\
16.4	9.12914\\
16.45	9.15272\\
16.5	9.17591\\
16.55	9.19874\\
16.6	9.22122\\
16.65	9.24337\\
16.7	9.2652\\
16.75	9.28672\\
16.8	9.30795\\
16.85	9.32888\\
16.9	9.34953\\
16.95	9.3699\\
17	9.38998\\
17.05	9.40979\\
17.1	9.42931\\
17.15	9.44855\\
17.2	9.46751\\
17.25	9.4862\\
17.3	9.50462\\
17.35	9.52277\\
17.4	9.54066\\
17.45	9.55828\\
17.5	9.57563\\
17.55	9.59272\\
17.6	9.60954\\
17.65	9.62606\\
17.7	9.64229\\
17.75	9.65821\\
17.8	9.67383\\
17.85	9.68914\\
17.9	9.70414\\
17.95	9.71882\\
18	9.73318\\
18.05	9.74722\\
18.1	9.76093\\
18.15	9.7743\\
18.2	9.78732\\
18.25	9.79997\\
18.3	9.81224\\
18.35	9.82412\\
18.4	9.83561\\
18.45	9.84671\\
18.5	9.85743\\
18.55	9.86778\\
18.6	9.87774\\
18.65	9.88731\\
18.7	9.89648\\
18.75	9.90527\\
18.8	9.91365\\
18.85	9.92161\\
18.9	9.92914\\
18.95	9.93624\\
19	9.94292\\
19.05	9.94918\\
19.1	9.95503\\
19.15	9.96048\\
19.2	9.96554\\
19.25	9.97021\\
19.3	9.97449\\
19.35	9.97839\\
19.4	9.98189\\
19.45	9.98499\\
19.5	9.98768\\
19.55	9.98997\\
19.6	9.99186\\
19.65	9.99338\\
19.7	9.99454\\
19.75	9.99536\\
19.8	9.99586\\
19.85	9.99605\\
19.9	9.99596\\
19.95	9.99561\\
20	9.995\\
20.05	9.99412\\
20.1	9.99298\\
20.15	9.99159\\
20.2	9.98993\\
20.25	9.988\\
20.3	9.98578\\
20.35	9.98328\\
20.4	9.98048\\
20.45	9.97735\\
20.5	9.9739\\
20.55	9.97011\\
20.6	9.966\\
20.65	9.96159\\
20.7	9.9569\\
20.75	9.95193\\
20.8	9.94674\\
20.85	9.94135\\
20.9	9.93579\\
20.95	9.9301\\
21	9.9243\\
21.05	9.91841\\
21.1	9.91244\\
21.15	9.90642\\
21.2	9.90036\\
21.25	9.8943\\
21.3	9.88824\\
21.35	9.88222\\
21.4	9.87622\\
21.45	9.87027\\
21.5	9.86438\\
21.55	9.85854\\
21.6	9.85277\\
21.65	9.8471\\
21.7	9.84152\\
21.75	9.83606\\
21.8	9.83072\\
21.85	9.82548\\
21.9	9.82037\\
21.95	9.81537\\
22	9.81049\\
22.05	9.80573\\
22.1	9.8011\\
22.15	9.7966\\
22.2	9.79223\\
22.25	9.78801\\
22.3	9.78395\\
22.35	9.78005\\
22.4	9.77633\\
22.45	9.7728\\
22.5	9.76946\\
22.55	9.76632\\
22.6	9.76339\\
22.65	9.76067\\
22.7	9.75815\\
22.75	9.75585\\
22.8	9.75373\\
22.85	9.75177\\
22.9	9.74995\\
22.95	9.74826\\
23	9.74668\\
23.05	9.74521\\
23.1	9.74385\\
23.15	9.74259\\
23.2	9.74144\\
23.25	9.7404\\
23.3	9.73948\\
23.35	9.73868\\
23.4	9.73799\\
23.45	9.73737\\
23.5	9.73682\\
23.55	9.73632\\
23.6	9.73585\\
23.65	9.73539\\
23.7	9.73492\\
23.75	9.73443\\
23.8	9.73391\\
23.85	9.73335\\
23.9	9.73275\\
23.95	9.73211\\
24	9.7314\\
};
\end{axis}
\node[above right, align=left, text=black, font=\scriptsize\linespread{0.8}\selectfont]
at (MyAxis.south west) {%
	\scriptsize{Experiment 2.1:} \\ \scriptsize{$w_c = 1$}};
\end{tikzpicture}%
		\end{subfigure}
		\begin{subfigure}[t]{0.5\textwidth}
			\input{tikzPlots/lehr_lat_lon_acc.tex}
		\end{subfigure}
		\begin{subfigure}[t]{0.5\textwidth}
%
%
\definecolor{mycolor1}{rgb}{0.00000,0.44700,0.74100}%
\begin{tikzpicture}[scale=0.75]

\begin{axis}[%
width=3.5in,
height=1in,
at={(0.758in,0.481in)},
scale only axis,
xmin=0,
xmax=24,
ymin=0,
ymax=12,
axis background/.style={fill=white},
ylabel={velocity in $\SI{}{\meter\per\second}$},
xlabel={time $t$ in $\SI{}{\second}$}
]
\addplot [color=mycolor1, forget plot]
  table[row sep=crcr]{%
0	10.0482\\
0.05	10.0518\\
0.1	10.0549\\
0.15	10.0575\\
0.2	10.0596\\
0.25	10.061\\
0.3	10.0617\\
0.35	10.0616\\
0.4	10.0607\\
0.45	10.0588\\
0.5	10.056\\
0.55	10.0522\\
0.6	10.0473\\
0.65	10.0413\\
0.7	10.0341\\
0.75	10.0258\\
0.8	10.0163\\
0.85	10.0055\\
0.9	9.99346\\
0.95	9.98015\\
1	9.96554\\
1.05	9.94961\\
1.1	9.93236\\
1.15	9.91378\\
1.2	9.89386\\
1.25	9.8726\\
1.3	9.84998\\
1.35	9.82603\\
1.4	9.80074\\
1.45	9.7741\\
1.5	9.74612\\
1.55	9.71681\\
1.6	9.68618\\
1.65	9.65423\\
1.7	9.62096\\
1.75	9.58639\\
1.8	9.55053\\
1.85	9.51337\\
1.9	9.47494\\
1.95	9.43526\\
2	9.39431\\
2.05	9.35212\\
2.1	9.3087\\
2.15	9.26406\\
2.2	9.21822\\
2.25	9.17118\\
2.3	9.12295\\
2.35	9.07357\\
2.4	9.02304\\
2.45	8.97136\\
2.5	8.91856\\
2.55	8.86466\\
2.6	8.80966\\
2.65	8.75358\\
2.7	8.69644\\
2.75	8.63826\\
2.8	8.57904\\
2.85	8.51881\\
2.9	8.45758\\
2.95	8.39538\\
3	8.33221\\
3.05	8.26809\\
3.1	8.20304\\
3.15	8.13709\\
3.2	8.07025\\
3.25	8.00253\\
3.3	7.93397\\
3.35	7.86458\\
3.4	7.79439\\
3.45	7.72341\\
3.5	7.65167\\
3.55	7.5792\\
3.6	7.50602\\
3.65	7.43214\\
3.7	7.35761\\
3.75	7.28245\\
3.8	7.20668\\
3.85	7.13033\\
3.9	7.05344\\
3.95	6.97604\\
4	6.89814\\
4.05	6.81979\\
4.1	6.74101\\
4.15	6.66185\\
4.2	6.58233\\
4.25	6.50248\\
4.3	6.42234\\
4.35	6.34195\\
4.4	6.26133\\
4.45	6.18053\\
4.5	6.09957\\
4.55	6.01851\\
4.6	5.93736\\
4.65	5.85617\\
4.7	5.77497\\
4.75	5.6938\\
4.8	5.61269\\
4.85	5.53169\\
4.9	5.45082\\
4.95	5.37012\\
5	5.28963\\
5.05	5.20937\\
5.1	5.12939\\
5.15	5.04972\\
5.2	4.97038\\
5.25	4.89141\\
5.3	4.81285\\
5.35	4.73471\\
5.4	4.65703\\
5.45	4.57984\\
5.5	4.50316\\
5.55	4.42702\\
5.6	4.35145\\
5.65	4.27646\\
5.7	4.20208\\
5.75	4.12833\\
5.8	4.05524\\
5.85	3.98281\\
5.9	3.91108\\
5.95	3.84005\\
6	3.76974\\
6.05	3.70017\\
6.1	3.63135\\
6.15	3.56329\\
6.2	3.49601\\
6.25	3.42951\\
6.3	3.36381\\
6.35	3.29892\\
6.4	3.23484\\
6.45	3.17158\\
6.5	3.10915\\
6.55	3.04755\\
6.6	2.98679\\
6.65	2.92686\\
6.7	2.86779\\
6.75	2.80956\\
6.8	2.75219\\
6.85	2.69566\\
6.9	2.63999\\
6.95	2.58518\\
7	2.53122\\
7.05	2.47811\\
7.1	2.42586\\
7.15	2.37446\\
7.2	2.32391\\
7.25	2.27421\\
7.3	2.22536\\
7.35	2.17735\\
7.4	2.13019\\
7.45	2.08386\\
7.5	2.03836\\
7.55	1.9937\\
7.6	1.94986\\
7.65	1.90684\\
7.7	1.86464\\
7.75	1.82325\\
7.8	1.78267\\
7.85	1.74289\\
7.9	1.7039\\
7.95	1.6657\\
8	1.62828\\
8.05	1.59163\\
8.1	1.55575\\
8.15	1.52063\\
8.2	1.48626\\
8.25	1.45263\\
8.3	1.41974\\
8.35	1.38756\\
8.4	1.3561\\
8.45	1.32535\\
8.5	1.29529\\
8.55	1.26591\\
8.6	1.23721\\
8.65	1.20916\\
8.7	1.18177\\
8.75	1.15502\\
8.8	1.12889\\
8.85	1.10338\\
8.9	1.07847\\
8.95	1.05415\\
9	1.03041\\
9.05	1.00723\\
9.1	0.984612\\
9.15	0.962533\\
9.2	0.940983\\
9.25	0.919949\\
9.3	0.899419\\
9.35	0.879381\\
9.4	0.859823\\
9.45	0.840733\\
9.5	0.822098\\
9.55	0.803907\\
9.6	0.786149\\
9.65	0.768812\\
9.7	0.751885\\
9.75	0.735358\\
9.8	0.719219\\
9.85	0.703458\\
9.9	0.688064\\
9.95	0.673029\\
10	0.658343\\
10.05	0.643995\\
10.1	0.629977\\
10.15	0.61628\\
10.2	0.602895\\
10.25	0.589814\\
10.3	0.57703\\
10.35	0.564533\\
10.4	0.552318\\
10.45	0.540376\\
10.5	0.5287\\
10.55	0.517283\\
10.6	0.50612\\
10.65	0.495204\\
10.7	0.484527\\
10.75	0.474086\\
10.8	0.463873\\
10.85	0.453883\\
10.9	0.444111\\
10.95	0.434551\\
11	0.425199\\
11.05	0.41605\\
11.1	0.407099\\
11.15	0.398341\\
11.2	0.389772\\
11.25	0.381387\\
11.3	0.373182\\
11.35	0.365155\\
11.4	0.357299\\
11.45	0.349612\\
11.5	0.342089\\
11.55	0.334728\\
11.6	0.327525\\
11.65	0.320475\\
11.7	0.313576\\
11.75	0.306825\\
11.8	0.300218\\
11.85	0.293753\\
11.9	0.287425\\
11.95	0.281233\\
12	0.275173\\
12.05	0.269243\\
12.1	0.263439\\
12.15	0.25776\\
12.2	0.252203\\
12.25	0.246764\\
12.3	0.241442\\
12.35	0.236234\\
12.4	0.231139\\
12.45	0.226152\\
12.5	0.221273\\
12.55	0.2165\\
12.6	0.211829\\
12.65	0.20726\\
12.7	0.202789\\
12.75	0.198416\\
12.8	0.194137\\
12.85	0.189952\\
12.9	0.185858\\
12.95	0.181854\\
13	0.177937\\
13.05	0.174107\\
13.1	0.170361\\
13.15	0.166697\\
13.2	0.163114\\
13.25	0.159611\\
13.3	0.156185\\
13.35	0.152835\\
13.4	0.149559\\
13.45	0.146357\\
13.5	0.143225\\
13.55	0.140164\\
13.6	0.137171\\
13.65	0.134245\\
13.7	0.131384\\
13.75	0.128588\\
13.8	0.125854\\
13.85	0.123181\\
13.9	0.120567\\
13.95	0.118012\\
14	0.115514\\
14.05	0.113072\\
14.1	0.110684\\
14.15	0.108348\\
14.2	0.106065\\
14.25	0.103831\\
14.3	0.101647\\
14.35	0.0995106\\
14.4	0.097421\\
14.45	0.0953768\\
14.5	0.093377\\
14.55	0.0914205\\
14.6	0.0895061\\
14.65	0.0876327\\
14.7	0.0857994\\
14.75	0.0840051\\
14.8	0.0822488\\
14.85	0.0805296\\
14.9	0.0788465\\
14.95	0.0771988\\
15	0.0755854\\
15.05	0.0740057\\
15.1	0.0724587\\
15.15	0.0709437\\
15.2	0.0694601\\
15.25	0.0680069\\
15.3	0.0665837\\
15.35	0.0651897\\
15.4	0.0638243\\
15.45	0.0624868\\
15.5	0.0611766\\
15.55	0.0598932\\
15.6	0.058636\\
15.65	0.0574045\\
15.7	0.0561982\\
15.75	0.0550165\\
15.8	0.053859\\
15.85	0.0527251\\
15.9	0.0516146\\
15.95	0.0505268\\
16	0.0494614\\
16.05	0.0484179\\
16.1	0.047396\\
16.15	0.0463951\\
16.2	0.045415\\
16.25	0.0444553\\
16.3	0.0435154\\
16.35	0.0425952\\
16.4	0.0424612\\
16.45	0.0437797\\
16.5	0.0471256\\
16.55	0.0529897\\
16.6	0.0617902\\
16.65	0.0738718\\
16.7	0.0895114\\
16.75	0.108924\\
16.8	0.132198\\
16.85	0.159381\\
16.9	0.190488\\
16.95	0.225503\\
17	0.264335\\
17.05	0.30688\\
17.1	0.353016\\
17.15	0.40261\\
17.2	0.455468\\
17.25	0.511398\\
17.3	0.570202\\
17.35	0.631687\\
17.4	0.695618\\
17.45	0.761769\\
17.5	0.82992\\
17.55	0.899862\\
17.6	0.97136\\
17.65	1.04419\\
17.7	1.11815\\
17.75	1.19305\\
17.8	1.26866\\
17.85	1.34482\\
17.9	1.42134\\
17.95	1.49805\\
18	1.57481\\
18.05	1.65145\\
18.1	1.72785\\
18.15	1.80388\\
18.2	1.87941\\
18.25	1.95432\\
18.3	2.02851\\
18.35	2.10188\\
18.4	2.17434\\
18.45	2.24579\\
18.5	2.31617\\
18.55	2.3854\\
18.6	2.45337\\
18.65	2.52\\
18.7	2.58521\\
18.75	2.64893\\
18.8	2.71109\\
18.85	2.77165\\
18.9	2.83055\\
18.95	2.88778\\
19	2.9433\\
19.05	2.99708\\
19.1	3.04912\\
19.15	3.09943\\
19.2	3.14801\\
19.25	3.19487\\
19.3	3.24004\\
19.35	3.28355\\
19.4	3.32545\\
19.45	3.36577\\
19.5	3.40458\\
19.55	3.44193\\
19.6	3.47789\\
19.65	3.51255\\
19.7	3.54597\\
19.75	3.57826\\
19.8	3.60952\\
19.85	3.63985\\
19.9	3.66936\\
19.95	3.69816\\
20	3.72639\\
20.05	3.75417\\
20.1	3.78164\\
20.15	3.80891\\
20.2	3.83613\\
20.25	3.86345\\
20.3	3.89099\\
20.35	3.91888\\
20.4	3.94726\\
20.45	3.97625\\
20.5	4.00598\\
20.55	4.03655\\
20.6	4.06809\\
20.65	4.1007\\
20.7	4.13447\\
20.75	4.16949\\
20.8	4.20583\\
20.85	4.24357\\
20.9	4.28276\\
20.95	4.32344\\
21	4.36563\\
21.05	4.40937\\
21.1	4.45465\\
21.15	4.50147\\
21.2	4.54984\\
21.25	4.59972\\
21.3	4.6511\\
21.35	4.70394\\
21.4	4.7582\\
21.45	4.81384\\
21.5	4.8708\\
21.55	4.92903\\
21.6	4.98848\\
21.65	5.04908\\
21.7	5.11078\\
21.75	5.17349\\
21.8	5.23716\\
21.85	5.30174\\
21.9	5.36713\\
21.95	5.43328\\
22	5.50012\\
22.05	5.5676\\
22.1	5.63565\\
22.15	5.7042\\
22.2	5.77319\\
22.25	5.84254\\
22.3	5.91218\\
22.35	5.98205\\
22.4	6.05208\\
22.45	6.1222\\
22.5	6.19237\\
22.55	6.26251\\
22.6	6.33256\\
22.65	6.40247\\
22.7	6.47218\\
22.75	6.54162\\
22.8	6.61075\\
22.85	6.67949\\
22.9	6.7478\\
22.95	6.81563\\
23	6.88289\\
23.05	6.94949\\
23.1	7.01537\\
23.15	7.08046\\
23.2	7.14474\\
23.25	7.20817\\
23.3	7.27074\\
23.35	7.33243\\
23.4	7.39327\\
23.45	7.4533\\
23.5	7.51255\\
23.55	7.57104\\
23.6	7.62877\\
23.65	7.68576\\
23.7	7.74199\\
23.75	7.79748\\
23.8	7.85221\\
23.85	7.90617\\
23.9	7.95935\\
23.95	8.01173\\
24	8.06333\\
};
\end{axis}
\node[above right, align=left, text=black, font=\scriptsize\linespread{0.8}\selectfont]
at (MyAxis.south west) {%
	\scriptsize{Experiment 2.2:} \\ \scriptsize{$w_c = 4$}};
\end{tikzpicture}%
		\end{subfigure}
		\begin{subfigure}[t]{0.5\textwidth}
			\input{tikzPlots/lehr_Lat_lon_acc_coop_costs.tex}
		\end{subfigure}
		\caption{T-junction Scenario, where two experiments are regarded. In Experiment 2.1, standard parameters are used as shown in Table \ref{tab:parameters}, where ego progress and courtesy costs are weighted equally. The first row depicts the temporal evolution of the scene and the second row shows velocity and accelerations. In Experiment 2.2, the same scenario is regarded, however, over courteous behavior is enforced with $w_c = 4$. The third row shows the corresponding velocity and accelerations. In Experiment 2.2, the courtesy costs are too high for merging in front of Vehicle 3. Therefore, only merging behind Vehicle 3 is possible. This emphasizes that the trade-off between social compliance and ego progress can be effectively regarded.}
		\label{fig:lehrOverview}
	\end{figure*}

	In addition, a more sophisticated scenario is evaluated where the corresponding T-junction is depicted in \figurename~\ref{fig:lehrOverview} and the target speed is $\SI{50}{\kilo\meter\per\hour}$ for all vehicles. In this scenario, the ego vehicle performs a left turn without right of way. First, the ego vehicle has to wait for the crossing Vehicle 1 approaching from the top with an initial speed of $\SI{30}{\kilo\meter\per\hour}$. Subsequently, Vehicle 2 and Vehicle 3 approach from the bottom with an initial speed of $\SI{30}{\kilo\meter\per\hour}$. In Experiment 2.1, where the parametrization shown in Table \ref{tab:parameters} is used, merging in front of Vehicle 2 would lead to high courtesy costs. Therefore, the ego vehicle merges behind it, in front of Vehicle 3. In order to make fast progress, the longitudinal acceleration is strongly increased at approximately $\SI{6}{\second}$. Subsequently, the longitudinal acceleration is slowly decreased in order to restrict the lateral acceleration in curved areas. Finally, the longitudinal acceleration is increased and decreased again to comfortably reach the appropriate distance to Vehicle 2. 
	In Experiment 2.2, the same scenario is regarded, however, over courteous behavior is enforced with an increased courtesy cost weighting $w_c = 4$. The corresponding velocity and accelerations are depicted in \figurename~\ref{fig:lehrOverview}. It is shown that the ego vehicle slows down for a long duration and in contrast to Experiment 2.1, does not even merge in front of Vehicle 3 in order to prevent high courtesy costs.
	This emphasizes that the trade-off between social compliant behavior and ego progress can be appropriately executed by the proposed approach. Hence, it is for example possible to adapt the courtesy to different traffic situations. 
	These results also indicate the suitability of the prediction concept in combination with the planning approach.
	
	In addition, we investigate the runtime of the framework.
	Table \ref{tab:runtime} summarizes the results measured for the evaluated scenarios as well as for each scenario with 10 additional other vehicles.
	The maximum time elapsed for combined environment processing and planning is $\SI{41}{\milli\second}$ where up to 13 other cars have to be regarded. 
	This clearly emphasizes the performance of the approach.
	
	{\renewcommand{\arraystretch}{1.0} 
		\begin{table}
			\caption{Runtime evaluation for environment processing and planning combined. In the scaled scenarios 10 vehicles where added to the scene. The calculation times are rounded to $\SI{}{\milli\second}$.}
			\label{tab:runtime}
			\begin{center}
				\begin{tabular}{c| l l}
					Scenario &Merging &T-junction \\
					\hline
					Max (as shown / scaled)&$\SI{37}{\milli\second}$/$\SI{41}{\milli\second}$& $\SI{32}{\milli\second}$ /$\SI{33}{\milli\second}$ \\
					Avg (as shown / scaled)&$\SI{17}{\milli\second}$/$\SI{19}{\milli\second}$& $\SI{13}{\milli\second}$/$\SI{15}{\milli\second}$\\
					\hline
				\end{tabular}
			\end{center}
			\vspace{-6mm}
		\end{table}
	}

	\section{Conclusion}
	In this work, a novel trajectory planning framework is presented which allows for social compliant behavior in intersection scenarios while considering available route and maneuver options of other vehicles.
	As a part of the framework, the local continuous optimization explicitly ensures that the obtained trajectories are comfortable and adhere to acceleration constraints.
	The evaluation showcases the effectiveness of the approach, where a long planning horizon for behavior options can be considered in real-time. Further, it is shown that the state of the art is effectively extended and outperformed.
	In future work, we would like to deploy the framework on the experimental vehicle of Ulm University and investigate its performance under real-world conditions.


	\bibliographystyle{../IEEEtranBST/IEEEtran.bst}

\end{document}